\documentclass[twoside]{article}

\usepackage[accepted]{aistats2026}
\usepackage{natbib} 
\usepackage{graphicx}
\usepackage{siunitx}
\usepackage{mathtools}
\usepackage{amsfonts}
\usepackage{booktabs}
\usepackage{multirow}
\usepackage{placeins}
\usepackage{subcaption}
\usepackage{amsmath}
\usepackage{xcolor}
\usepackage{tikz}
\usetikzlibrary{positioning, fit, arrows.meta, backgrounds, calc}

\usepackage{subcaption} 
\newcommand{\Reals}{\mathbb{R}}
\newcommand{\channel}{C}

\newcommand{\treeheight}{y}
\newcommand{\quantilesnum}{N}
\newcommand{\quantilesind}{n}
\newcommand{\quantile}{Q}
\newcommand{\timesteps}{T}

\newcommand{\height}{H}
\newcommand{\heightind}{h}
\newcommand{\width}{W}
\newcommand{\widthind}{w}

\newcommand{\inputspace}{\mathcal{X}}
\newcommand{\InputSpace}{\mathbb{R}^{\timesteps \times \channel \times \height \times \width}}

\newcommand{\LabelSpace}{\mathbb{R}^{\height \times \width}}
\newcommand{\outputspace}{\mathcal{Y}}
\newcommand{\inputimage}{X}
\newcommand{\outputimage}{Y}
\newcommand{\sparselabelimage}{\mathbb{\labelimage}}
\newcommand{\track}{t}
\newcommand{\tracknum}{I}
\newcommand{\trackind}{i}
\usepackage{tikz}
\usetikzlibrary{shapes,arrows,shadows}

\newcommand*{\mycolorbox}[1]{%
\tikzstyle{mybox} = [draw=black, rectangle, inner sep=1pt, inner ysep=2pt, inner xsep=2pt, fill=white]
\tikzstyle{fancytitle} = [fill=white, text=black]
\begin{tikzpicture}
\node [mybox] (box){%
     #1
};
\end{tikzpicture}%
}
\newcommand{\labelimage}{Y}
\newcommand{\gt}{y}
\newcommand{\pred}{\hat{y}}
\newcommand{\empiricalcoverage}{\mathrm{EC}}
\newcommand{\datasetlength}{M}
\newcommand{\datasetind}{m}
\newcommand{\SparseLabelSpace}{\outputspace_{\mathrm{sparse}}}
\DeclareMathOperator*{\argmin}{arg\,min}

\begin{document}



\twocolumn[

\runningtitle{Canopy Tree Height Estimation Using Quantile Regression}
\aistatstitle{Canopy Tree Height Estimation Using Quantile Regression: Modeling and Evaluating Uncertainty in Remote Sensing}

\aistatsauthor{ Karsten Schrödter \And Jan Pauls \And  Fabian Gieseke }

\aistatsaddress{ University of Münster \And  University of Münster \And University of Münster \\ University of Copenhagen } ]

\begin{abstract}
Accurate tree height estimation is vital
for ecological monitoring and biomass assessment. We apply quantile regression to existing tree height estimation models based on satellite data to incorporate uncertainty quantification. Most current approaches for tree height estimation rely on point predictions, which limits their applicability in risk-sensitive scenarios. In this work, we show that, with minor modifications of a given prediction head, existing models can be adapted to provide statistically calibrated uncertainty estimates via quantile regression.
Furthermore, we demonstrate how our results correlate with known challenges in remote sensing (e.g., terrain complexity, vegetation heterogeneity), indicating that the model is less confident in more challenging conditions. 
\end{abstract}

\section{Introduction}

Monitoring forests is an important component of assessing the effort required to comply with the Paris Climate Agreement.\footnote{For instance, on a global scale, forests absorb $3.5 \pm \SI{0.4}{\peta\gram}$ of carbon per year~\citep{pan24}.} In recent years, the availability of satellite remote sensing data at high resolution and at a global scale as well as advances in machine learning have paved the way for an automated forest monitoring. A typical building block for analyses in this domain is the estimation of (canopy) tree height from remote sensing data, which then serves as a proxy for follow-up analyses such as so-called above-ground biomass estimation \citep{schwartzFORMSForestMultiple2023a} and, thus, quantification of stored carbon. 

\begin{figure}[t]
    \centering
    \begin{subfigure}[t]{0.29\columnwidth}
        \centering      \mycolorbox{\includegraphics[width=\linewidth]{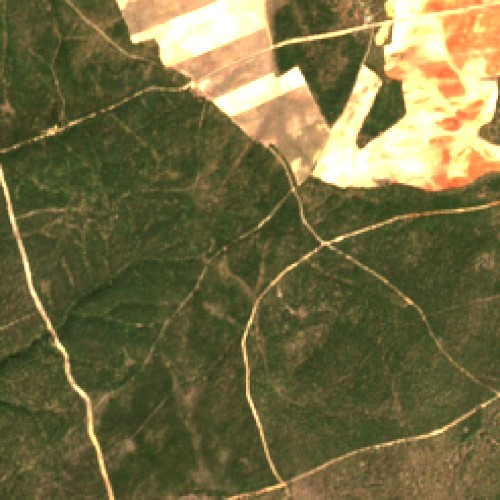}}
        \caption{}
    \end{subfigure}
    \hfill
    \begin{subfigure}[t]{0.29\columnwidth}
        \centering
    \mycolorbox{\includegraphics[width=\linewidth]{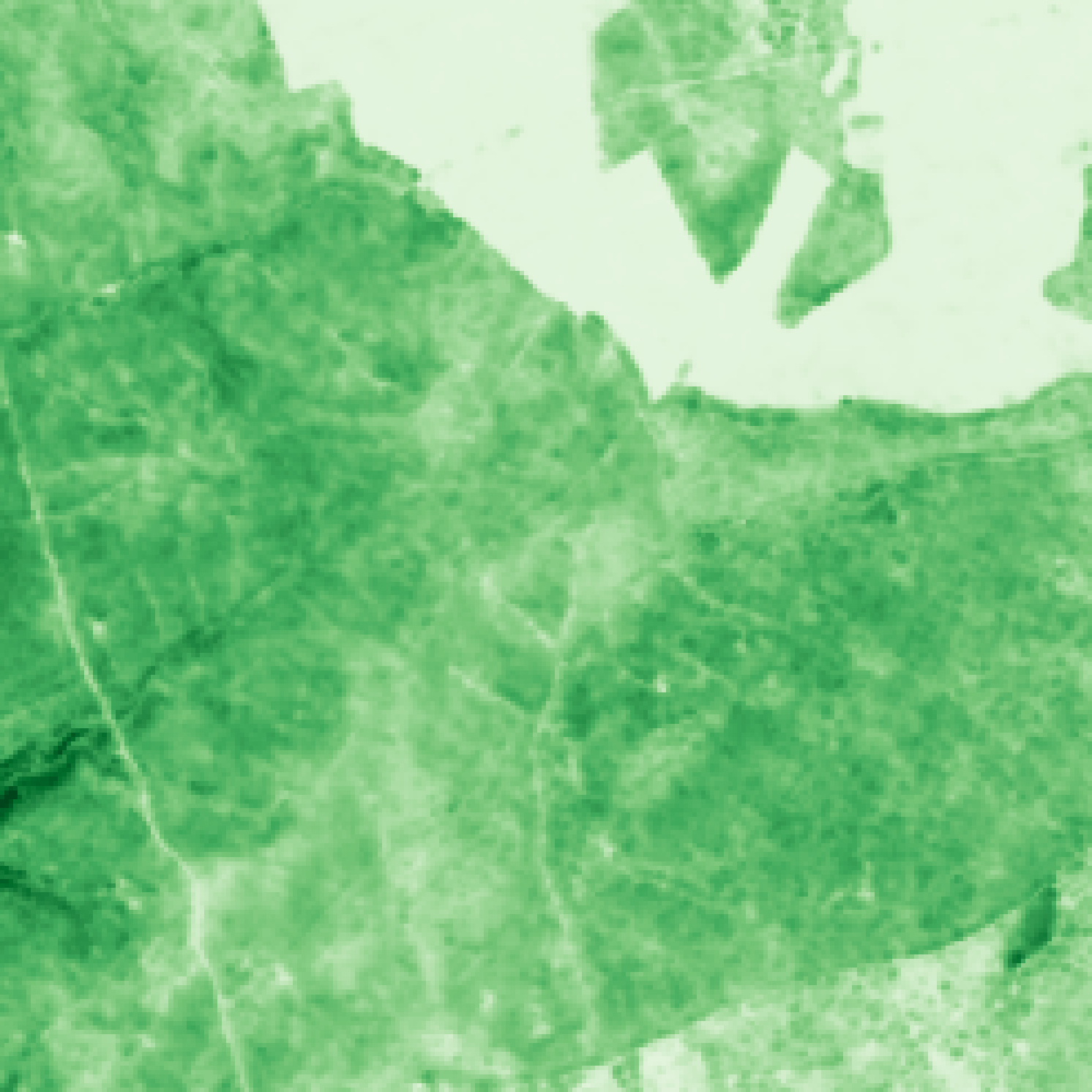}}
    \caption{}
    \end{subfigure}
    \hfill
    \begin{subfigure}[t]{0.29\columnwidth}
        \centering
     \mycolorbox{\includegraphics[width=\linewidth]{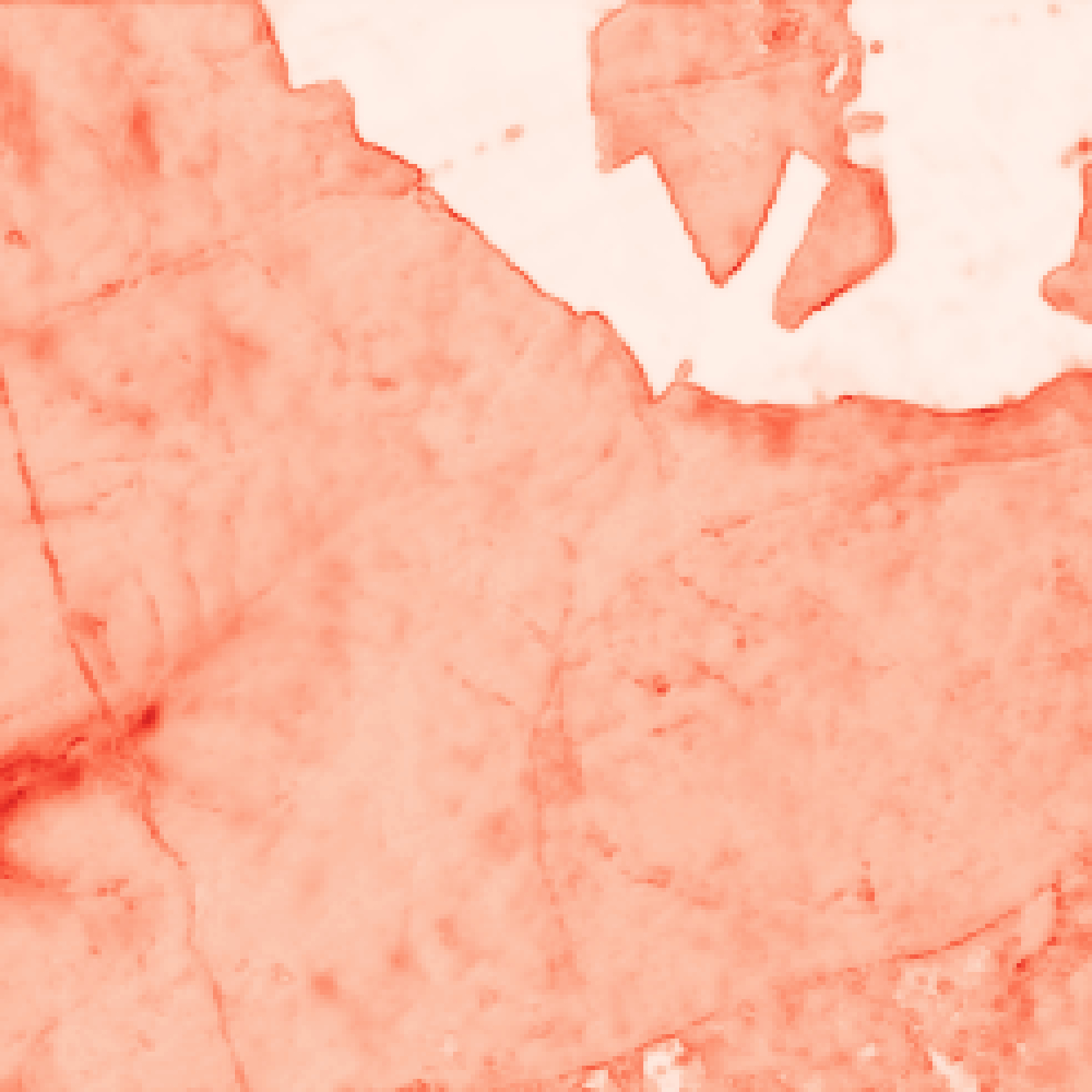}}
     \caption{}
    \end{subfigure}
    \hfill
    \vskip-0.1cm
    \caption{Tree height estimation: An input image (a) can be used to produce a tree height map (b) using a tree height model (intensity of green increases with higher tree height estimation). Most studies focus on point estimates. In Figure~(c), associated uncertainty estimates are provided (intensity of red increases with higher uncertainty). It can be seen that the model is less confident at forest borders and along cutoff stripes.}
    \label{fig:motivation}
    \vspace{-0.2cm}
\end{figure}

An example of such a tree height estimation scenario is shown in Figure~\ref{fig:motivation}. Here, the input satellite data (a) are used to obtain an estimate (b) for the tree height in the corresponding area.
Most recent studies have focused on using point-estimators to produce canopy height maps from country-to-global scale~\citep{schwartz2022high,tolanVeryHighResolution2024,pauls2024,pauls2025, potapovMappingGlobalForest2021}. However, valid uncertainty estimation for canopy height maps is rare, with \cite{Lang2023} among the few exceptions. Figure~\ref{fig:motivation}~(c) shows an uncertainty map for the height estimations given in Figure~(b). It can be seen, for instance, that the model is less confident at forest borders.
Uncertainty quantification is essential for decision- and policy-making, where worst-case scenarios must be incorporated into climate simulations and general risk assessments~\citep{Smith2011}. In forest monitoring, quantifying uncertainty in canopy-height estimates enables its propagation through non-linear downstream models of above-ground biomass and carbon storage, underpinning robust derived products such as $\mathrm{CO_2}$ certificates.

In this work, we use quantile regression to incorporate uncertainty quantification into existing pre-trained tree height models, evaluate their effectiveness using recently proposed architectures, and analyze the resulting uncertainty maps on real-world data. On held-out validation sets, we show that the predicted uncertainties are well calibrated and we characterize recurrent conditions that systematically lead to elevated uncertainty. We expect our findings to advance the use of uncertainty quantification for this task, and to pave the way for a more reliable use of height maps for various downstream tasks, including above-ground biomass estimation and carbon accounting.

\section{Background}
We start by providing details about canopy tree height estimation along with uncertainty quantification for deep learning models.

\subsection{Canopy Height Estimation}
\label{sec:background_canopy}

Above-ground biomass is essential for carbon storage estimation and forest monitoring. However, not much ground-truth data is directly available, so tree height as an approximator is frequently used \citep{schwartzFORMSForestMultiple2023a}. Early approaches relied on manual field sampling and spatial interpolation, which were both costly and limited in coverage. The advent of free and open satellite imagery from missions such as Landsat and Sentinel-2 enabled tree height estimation at high resolution and continental to global scales.

Modern canopy height estimation leverages machine learning models trained on satellite imagery, such as the optical images from Sentinel-2, with ground truth labels derived from spaceborne LiDAR missions, like the Global Ecosystem Dynamics Investigation (GEDI) \citep{dubayahGEDILaunchesNew2022}.
GEDI and ICESat-2 \citep{neumann2019ice} missions provide sparse but accurate tree height measurements on a global scale. GEDI, a full-waveform LiDAR instrument aboard the International Station (ISS), records footprints with a beam diameter of approximately $\SI{25}{\meter}$, spaced $\SI{60}{\meter}$ apart along-track, across 8 parallel tracks. Geolocation is determined via GPS, star trackers, and instrument orientation data, achieving a reported 1-sigma accuracy of approximately $\SI{10.2}{\meter}$, though accuracy varies across individual tracks.

The spatial resolution of canopy height products has steadily improved alongside advances in satellite imagery availability. Early studies build on $\SI{30}{\meter}$ resolution data from Landsat (available for 30+ years). More recent work exploits $\SI{10}{\meter}$ resolution Sentinel-2 imagery (openly available since 2015), while commercial satellites provide $\SI{3}{\meter}$ daily coverage and sub-meter resolution for targeted regions. Current studies span multiple spatial scales: regional \citep{rolf2024contrasting}, national \citep{schwartz2022high,su2025canopy}, continental \citep{pauls2025,liu2023overlooked}, and global \citep{tolanVeryHighResolution2024,Lang2023,pauls2024,potapovMappingGlobalForest2021}, with resolutions ranging from $\SI{30}{\meter}$ to sub-meter scales.

Despite continuous improvements in point prediction accuracy, uncertainty quantification remains largely unaddressed. To our knowledge, \cite{Lang2023} is the only large-scale study that provides uncertainty estimates.
This represents a critical gap, as uncertainty estimates are essential for downstream applications: they enable reliable risk assessment in carbon offset programs, inform targeted allocation of resources for field campaigns, and allow propagation of prediction uncertainty through non-linear biomass and carbon storage models \citep{yambayamba2025tree,lin2023scaling,bian2023uncertainty,candelas2025assessing}. Without calibrated uncertainty estimates, canopy height predictions cannot be safely integrated into climate policy decisions that require worst-case scenario analysis \citep{besic2025remote}.

\subsection{Uncertainty in Deep Learning}

Methods for uncertainty quantification can roughly be put into 4 categories, see \cite{Gawlikowski2023}: Bayesian approaches, ensemble methods, test-time augmentation and single deterministic methods. 
Bayesian models put distributions on model parameters, leading to posterior-predictive distributions. 
Ensembles aim to infer uncertainty having multiple models that are initialized independently. 
Bayesian and ensemble approaches are highly compute-intense, that is why we do not apply them for canopy height estimation. 
Test-time augmentation methods apply multiple data augmentation techniques to input data at prediction time, often used in healthcare applications. This is challenging in the context of canopy height estimation, as there are few well established augmentations that keep canopy height labels unchanged. 
Single deterministic methods summarize all methods, where the uncertainty is generated in a single forward pass of a deterministic network. An example is Gaussian regression, where the output is modeled to follow a Gaussian distribution by estimating mean and variance and is trained to minimize the negative log-likelihood, a method that was introduced latest in 1994, see \cite{Nix1994}. 

By now, \cite{Lang2023} proposed the only approach for incorporating uncertainty quantification into a canopy height model by using an ensemble of Gaussian regression models. 
More single deterministic methods are tested in other biological applications of image-to-image regression tasks. 
Regression on the magnitude of residuals aims to predict the model's error for each prediction. Moreover, regression can be re-framed as a classification task and a softmax distribution can be used for uncertainty quantification. 
Another well-known technique is quantile regression, where an asymmetric loss function is used to predict conditional quantiles. All these methods are tested in \cite{Angelopoulos22} for different image-to-image regression tasks, with the finding that quantile regression performs best among these methods. There are theoretical results that guarantee good behaviour of quantile regression under mild assumptions, see \cite{Stein11}. Quantile regression is reported to show good results on image-to-image tasks, so we decided to apply it to canopy height estimation. 

\section{Approach}
Let $\inputspace$ and $\outputspace$ be the input and output space of our supervised learning task. An input datum consists of $\timesteps$ images of height $\height$ and width $\width$, each having $\channel$ channels. 
Therefore, the input space will be denoted as $\inputspace = \InputSpace$. An output consists of $\quantilesnum$ images of the same size, while a label consists of a single image of the same size. 
Thus, the label space will be denoted by $\outputspace = \LabelSpace$. Each output image aims to train a quantile canopy height model $f: \mathcal{X} \to \mathcal{Y}$ by minimizing some loss on training data 
$ \mathcal{D} = ((\inputimage_1, \outputimage_1),\dots, (\inputimage_\datasetlength, \outputimage_\datasetlength)) \in (\inputspace \times \outputspace)^\datasetlength $.
Note that since the labels in $\mathcal{D}$ are generated from GEDI measurements, it contains sparse information, see the right part of Figure ~\ref{fig:noisy_labels}. Having no information on a label $\labelimage \in \mathcal{Y}$ at an index $(\heightind,\widthind)$ is encoded by $\labelimage_{\heightind,\widthind} =0$.
For a label $\labelimage \in \mathcal{Y}$ we define its sparse notation by 
$$\sparselabelimage \coloneq \{(\heightind, \widthind, \treeheight) \mid 1 \le \heightind \le \height,  1 \le \widthind \le \width, \labelimage_{\heightind,\widthind} = \treeheight > 0\},$$
where $(\heightind, \widthind, \treeheight) \in \sparselabelimage$ reads as the ground truth at pixel $(\heightind,\widthind)$ is $\treeheight$. 
We denote the space of sparse labels by $\SparseLabelSpace$. The sparse notation is introduced to facilitate the notation of the loss functions in our context. 

\subsection{Pinball Loss}

The basic idea behind quantile regression is to use the pinball loss, which penalizes over- and underprediction asymmetrically. Let $\tau \in (0,1)$ be a target quantile then the pinball loss $\mathcal{L}_{\tau}: \Reals \times \Reals \to \Reals^+ \coloneq [0,\infty)$ is defined by 
\begin{equation}
    \mathcal{L}_{\tau}(\gt, \pred) \coloneq 
\left\{
\begin{array}{rclc}
\tau ~ &\cdot& (\gt - \pred) & \text{if } \pred \le \gt \\
(1 - \tau) &\cdot& (\pred - \gt) & \text{if } \pred > \gt
\end{array}\right.,
\end{equation} 
where $\gt \in \Reals$ is the ground truth value and $\pred \in \Reals$ the prediction. For example, if $\tau = 0.2$, note that underprediction ($\pred \le \gt$) is penalized with a small factor of $\tau = 0.2$, while overshooting ($\pred > \gt$) is penalized with a higher factor of $1-\tau = 0.8$.  

Let $P$ be a distribution on $\inputspace \times \Reals$ and define the conditional $\tau$-quantile of $x \in \inputspace$ as  
$$
\quantile_\tau(x) \coloneq \inf\{ q \in \Reals \mid  P(y \le q \mid x) \ge \tau \}.$$ 
Then it is well-known, see e.g. \cite{Stein11}, that the function $\quantile_\tau$ sending $x \in \inputspace$ to $\quantile_\tau(x)$ is the minimizer among all measurable functions $f: \inputspace \to \mathbb{R}$ for the risk of the pinball loss, i.e.
$$ \quantile_\tau = \argmin\limits_{f: \inputspace \to \mathbb{R}} \int\limits_{\inputspace \times \mathbb{R}} \mathcal{L}_\tau(y, f(x)) dP(x,y).$$

\subsection{Pixelwise Quantile Regression}

In the context of image-to-image regression, the loss for pixelwise quantile regression averages the pinball loss at every pixel, as in \cite{Angelopoulos22}. In our context, we only have sparse labels and thus only average over the pixels with label. More precisely, let $\outputimage \in \Reals^{\height \times \width}$ be a prediction and denote the ground truth value in sparse notation by $\sparselabelimage \in \SparseLabelSpace$. 
% $ \sparselabelimage = \{ (\heightind, \widthind, \treeheight) \mid 1 \le \heightind \le \height, ~ 1 \le \widthind \le \width, \treeheight \in \Reals^+ \} $, 
Then the sparse pixelwise pinball loss is defined as 
\begin{equation}
\mathcal{L}_{\tau}(\sparselabelimage,\outputimage) \coloneq \frac{1}{\#\sparselabelimage} \sum\limits_{(\heightind,\widthind,\treeheight) \in \sparselabelimage} \mathcal{L}_\tau(\treeheight,\outputimage_{\heightind,\widthind}) \in \Reals^+.\end{equation}
With our model, we train multiple output images simultaneously on various quantiles $\tau_1, \dots, \tau_\quantilesnum \in (0,1)$. The final loss function is the average of all pinball losses of the output images, more precisely, denote the $\quantilesind$-th output image of the model by $\outputimage^{(\quantilesind)} \in \Reals^{\height \times \width} $ and consider a sparse ground truth label $\sparselabelimage \in \SparseLabelSpace$. Then the loss function $\mathcal{L}_{\bar{\tau}}$ is 
\begin{equation}
      \mathcal{L}_{\bar{\tau}}(\sparselabelimage, \mathbf{\outputimage}) \coloneq \frac{1}{ \quantilesnum } \sum\limits_{\quantilesind = 1}^\quantilesnum   \mathcal{L}_{\tau_\quantilesind}(\sparselabelimage, \outputimage^{(\quantilesind)}) \in \Reals^+, 
\end{equation}  
where $\bar{\tau} = (\tau_1, \dots, \tau_\quantilesnum) \in (0,1)^\quantilesnum$ is the vector of quantiles and $\mathbf{Y} = (\outputimage^{(1)},\dots, \outputimage^{(\quantilesnum)}) \in \Reals^{\height \times \width \times \quantilesnum}$ the concatenated output. 

\subsection{Shift-Resilient Loss}
\label{sec:shift_resilient_loss}

As in \cite{pauls2024}, we use a shifted variant of our loss. GEDI labels can be divided into tracks, i.e. into measurements that are taken on the same overflight path. The basic observation for the shift-resilient loss is that geolocation errors tend to be constant along tracks. Therefore, the idea is to allow the model to shift the labels within a track by a small offset, if this decrease the overall loss of the track. 

More precisely, a sparse label $\sparselabelimage \in \SparseLabelSpace$  can be partitioned into GEDI tracks 
that can be describe by $ \track_1,\dots,\track_{\tracknum} \in \SparseLabelSpace$ such that 
$$ \track_\trackind \subset \sparselabelimage, ~ \bigcup\limits_{\trackind=1}^{\tracknum} t_\trackind = \sparselabelimage, ~ t_\trackind \cap t_j = \emptyset $$
for $1 \le \trackind \ne j \le \tracknum$. All points of the same track are approximately on one line.
For a track $t \subset \sparselabelimage$ and a shifting direction $\delta = (\delta_\heightind,\delta_\widthind) \in \mathbb{Z}^2$ we define the shifted track $\track(\delta)$ by $$\track(\delta) \coloneq \{ (\heightind + \delta_\heightind, \widthind + \delta_\widthind, \treeheight) \mid (\heightind, \widthind, \treeheight) \in \track \}.$$
The shifted (sh) pixelwise quantile regression loss for a track $\track$, a quantile vector $\bar{\tau} = (\tau_1, \dots, \tau_\quantilesnum) \in (0,1)^\quantilesnum$ and a multi-image output $\mathbf{Y} \in \Reals^{\height \times \width \times \quantilesnum}$ is given by 
\begin{equation} \mathcal{L}_{\bar{\tau}}^{\mathrm{sh}}(\track,\mathbf{Y}) \coloneq \min\limits_{\delta \in \{-1,0,1\}^2} \mathcal{L}_{\bar{\tau}}(\track(\delta),\mathbf{Y}).
\end{equation}
In particular, the track is allowed to be shifted by at most one pixel in each direction to find the best fitting predictions. The shifted pixelwise regression loss for a sparse label $\sparselabelimage$ containing of tracks $ \track_1,\dots,\track_{\tracknum} \in \SparseLabelSpace$ is given by the average of all track losses, i.e.
\begin{equation}
\mathcal{L}_{\bar{\tau}}^{\mathrm{sh}}(\sparselabelimage,\mathbf{Y}) \coloneq \frac{1}{\tracknum}\sum\limits_{\trackind = 1
}^{\tracknum}\mathcal{L}_{\bar{\tau}}^{\mathrm{sh}}(\track_\trackind,\mathbf{Y}).
\end{equation}

\section{Experiments}
\label{sec:experiments}

We incorporate quantile regression into an existing tree canopy height model by adding an uncertainty head to the model. After fine-tuning of the model, we evaluate the point-estimator and the uncertainty estimates using metrics and qualitative comparisons to the model from \citet{Lang2023}.

\subsection{Experimental Setup}
To evaluate our approach, we perform several experiments testing the applicability to tree canopy height prediction on continental-scale. We use the model from \citet{pauls2025} and use an additional uncertainty head, which is trained using shifted pixelwise pinball loss. We fine-tune both the original and the uncertainty head on the same dataset for a quarter of the pretraining epochs and evaluate the model on a test dataset with $250$ samples from $2020$, containing approximately $49$k GEDI labels. We compare our results to \citet{Lang2023} and show analysis on uncertainty effects near forest borders and mountainous terrain. Additionally, we performed an ablation study on exchanging quantile regression by either Gaussian or Log-Gaussian regression, which aim to maximize the average likelihood of uncertainty predictions using a Gaussian, resp. Log-Gaussian, distribution. Gaussian regression is also used in \citet{Lang2023} for every ensemble member. 

In the following, we shortly describe the model, our fine-tuning strategy and the evaluation metrics.

\subsubsection{Data and Architecture}

\citet{pauls2025} propose to use a Unet model that is adopted for three-dimensional input, i.e. processing not just one satellite image, but rather a time-series. More precisely, they take a time-series of $\timesteps = 12$ monthly Sentinel-2 images of size $\height = \width = 256$ and concatenate a yearly composite of Sentinel-1 radar data channel-wise, resulting in $C = 16$ channels. Their ground-truth data stems from GEDI measurements that are filtered using several quality criteria to limit the influence from noisy labels.

We use the pretrained head and fine-tune it to predict a $50\%$-quantile, usable as point-estimator. Additionally, we aim to train the model to predict $10$ more quantiles, namely: $$\bar{\tau} = (0.05,0.1,0.15,0.2,0.25,0.75,0.8,0.85,0.9,0.95)$$ 
In total, the model predicts $\quantilesnum = 11$ channels. 
We adapt the architecture from \cite{pauls2025} by adding a new head, see Appendix \ref{app:architecture} for details. In the original architecture the head consisted of one $1 \times 1$ convolution that reduces the channels from 64 to 1. We build a second path and append an uncertainty head to the same 64 input channels. The head consists of two convolution blocks followed by a similar $1 \times 1$ convolution to reduce the channels from 64 to 10. One convolution block consists of a 2D convolution with a kernel of size $3 \times 3$ and same padding that do not alter the number of channels.

\subsubsection{Training Setting}

The model from \citet{pauls2025} was pretrained on Huber loss using the Adam optimizer \citep{kingma2014adam} with an initial learning rate $0.001$, weight decay of $0.01$ and gradient clipping at $1.0$. Training was done for a total of $400{,}000$ iterations with batch size 16 and a linear learning rate scheduler with $10\%$ warmup iterations was used. Our fine-tuning follows a similar procedure, however the shifted pixelwise pinball loss is used and we only fine-tune for 2 epochs using a batch size of $5$ and freeze everything except the two heads (point-estimator and uncertainty head).

\subsubsection{Uncertainty Quantification Evaluation}

We evaluate our predictions using several metrics that measure prediction intervals width, coverage and more. Here we introduce these concepts:

Let $\mathcal{T} = ((X_1,\sparselabelimage_1), \dots, (X_\datasetlength, \sparselabelimage_\datasetlength)) \in (\inputspace \times \outputspace)^{\datasetlength}$ be a test dataset, $\tau \in (0,1)$ be the desired quantile and $f_\tau: \inputspace \to \outputspace$ be a function -- e.g. a trained neural network -- to predict the $\tau$-conditional quantile.  
We evaluate the predictions for a quantile $\tau \in (0,1)$ by calculating the Empirical Coverage (EC), defined by 
\begin{equation}
    \empiricalcoverage_\tau =  \frac{\sum\limits_{\datasetind = 1}^{\datasetlength} \#\{ (\heightind,\widthind,\treeheight) \in \sparselabelimage_{\datasetind} \mid f_\tau(\inputimage_\datasetind)_{\heightind,\widthind} \ge \treeheight \}}{\sum\limits_{\datasetind = 1}^{\datasetlength} \# \sparselabelimage_{\datasetind}}, 
\end{equation}
which is essentially the proportion of labels that are overestimated by $f_\tau$ among all labeled points. 

Furthermore, we form prediction intervals (PI) for uncertainty level $\alpha \in (0,1)$ by predicting lower and upper quantiles $\tau_l$ and $\tau_h$, defined by  
$$ \tau_l \coloneq 0.5 - \alpha/2 ~\mathrm{and}~ \tau_h \coloneq 0.5 + \alpha/2,$$
and span an interval around the corresponding predictions, i.e. $$PI_\alpha(\inputimage) = [f_{\tau_l}(\inputimage),f_{\tau_h}(\inputimage)]$$ for an input image $\inputimage \in \inputspace$. Furthermore, we set the Prediction Interval Width (PIW) to 
\begin{equation}
    \mathrm{PIW}_\alpha(X) \coloneq f_{\tau_h}(\inputimage) - f_{\tau_l}(\inputimage) \in \LabelSpace
\end{equation}
and use established uncertainty metrics such as Mean Prediction Interval Width (MPIW) and Prediction Interval Coverage Probability (PICP), defined as 
\begin{align}
    &\mathrm{MPIW}_\alpha \coloneq \frac{\sum\limits_{\datasetind = 1}^{\datasetlength} \sum\limits_{(\heightind,\widthind, \treeheight) \in \sparselabelimage_\datasetind}
    % f_{\tau_h}(\inputimage_\datasetind) - f_{\tau_l}(\inputimage_\datasetind) }
    \mathrm{PIW}_\alpha(\inputimage_\datasetind)_{\heightind, \widthind} }
    {\sum\limits_{\datasetind = 1}^{\datasetlength} \# \sparselabelimage_{\datasetind}}, \\
    &\mathrm{PICP}_\alpha \coloneq \frac{\sum\limits_{\datasetind =  1}^{\datasetlength} \#\{ (\heightind,\widthind,\treeheight) \in \sparselabelimage_{\datasetind} \mid \treeheight \in  \mathrm{PI}_\alpha(\inputimage_\datasetind)\}}{\sum\limits_{\datasetind = 1}^{\datasetlength} \# \sparselabelimage_{\datasetind}}.
\end{align}

\subsection{Results}
As our approach serves as a mere add-on, we want to ensure that the model's point-prediction is not worsened after our architectural changes and fine-tuning. Section~\ref{sec:res_one_point} shows experiments comparing both stages and \citeauthor{Lang2023}'s prediction to the available ground-truth measurements. Sections~\ref{sec:uncertainty_results} - \ref{sec:res_high_pred_high_uncertainty}\ analyze the Prediction Interval Width from different perspectives, Sections~\ref{sec:res_forest_borders} and \ref{sec:res_mountains} examine the uncertainty at forest borders and mountainous areas, and Section~\ref{sec:res_qualitative} compares the visual prediction quality. Appendix \ref{app:ablation_gaussian} presents the results of the ablation study comparing quantile regression with Gaussian and Log-Gaussian regression. Quantile regression outperformed both alternatives in terms of point estimation and uncertainty quantification. We therefore focus on quantile regression in the following.

Furthermore, we refer to Appendix \ref{app:computation_overhead} for an analysis of the computational overhead of our architecture compared to the model from \citet{pauls2025}.

\subsubsection{Point Estimation Comparison}
\label{sec:res_one_point}
The point-estimator results of the three models (\citet{Lang2023}, \cite{pauls2025}, and Ours)  
are summarized in Table~\ref{tab:one_point_results}. 
The results are in line with findings in \citet{pauls2025} that the model from \citeauthor{pauls2025} yields better metrics than \citeauthor{Lang2023}. 
Note that fine-tuning the prediction head using the pinball loss for $\tau = 0.5$, i.e. essentially using Mean Absolute Error loss, did not change metrics drastically compared to \citeauthor{pauls2025}. However, it is worth mentioning that the prediction of \citeauthor{pauls2025}'s model, which is trained using the Huber loss, is above the ground truth GEDI value in about $58\%$ of the cases, while the fine-tuned $50\%$-quantile prediction is well-calibrated. 
Note that the $\mathrm{EC}$ values for \citeauthor{pauls2025} and \citeauthor{Lang2023} are only given as a reference, since both models are not trained to predict a $50\%$-quantile, but rather a mean in the case of \citeauthor{Lang2023}, or a combination of mean and median (\citeauthor{pauls2025}). 

\begin{table}[h]
\caption{Results on point-estimators, measured using Mean Squared Error (MSE, \si{\meter\squared}), Mean Absolute Error (MAE, \si{\meter}), $R^2$ and Empirical Coverage (EC).}  \label{tab:one_point_results}
\begin{center}
\begin{tabular}{l|cccc}
\toprule
Model & MSE $\downarrow$ & MAE $\downarrow$ & $R^2$ $\uparrow$ & $EC$ \\
\midrule
\citeauthor{Lang2023} & 38.32 & 4.25 & 0.55 & 0.47 \\
\citeauthor{pauls2025} & 20.64 & 1.90 & 0.76 & 0.58 \\
Ours & 20.81 & 1.90 & 0.76 & 0.50 \\
\bottomrule
\end{tabular}
\end{center}
\end{table}

\subsubsection{Uncertainty Quantification Evaluation}
\label{sec:uncertainty_results}

\begin{table*}[t]
\caption{Comparison of Mean Prediction Interval Width (MPIW) and Prediction Interval Coverage Probability (PICP) for multiple uncertainty levels $\alpha$ for \citeauthor{Lang2023}'s model and our model.} \label{table:uncertainty_results}
\begin{center}
\begin{tabular}{r|ccccc|ccccc}
\toprule
Metric & \multicolumn{5}{c|}{MPIW} & \multicolumn{5}{c}{PICP} \\
 $\alpha$ & $0.5$ & $0.6$ & $0.7$ & $0.8$ & $0.9$ & $0.5$ & $0.6$ & $0.7$ & $0.8$ & $0.9$ \\
\midrule
Lang et. al. & 5.54 & 6.87 & 8.39 & 10.22 & 12.76 & 0.37 & 0.45 & 0.52 & 0.58 & 0.64 \\
Ours & 2.48 & 3.07 & 3.79 & 4.77 & 6.47 & 0.48 & 0.58 & 0.67 & 0.78 & 0.88 \\
\bottomrule
\end{tabular}
\end{center}
\end{table*}

\begin{figure}[t]
\centering
\includegraphics[width=\columnwidth]{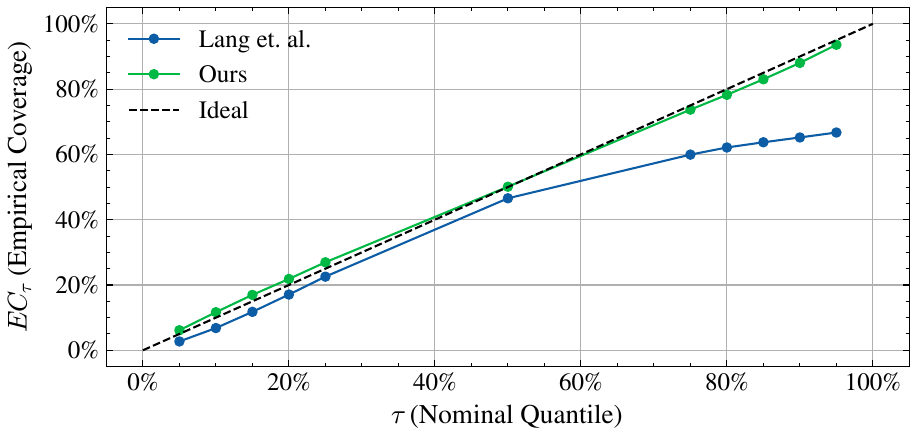}
\caption{Empirical Coverage vs Nominal Quantile for \citeauthor{Lang2023}'s and our model.}
\label{fig:empirical_coverage}
\end{figure}

Comparing the EC across all trained quantile-levels, depicted in Figure ~\ref{fig:empirical_coverage}, reveals that \cite{Lang2023} slightly underestimates the smaller quantiles, where our model's quantile predictions are consistently a bit too high. Noticeably, on higher quantiles, our model achieves similar performance, but underestimates the quantiles, where \cite{Lang2023} is covering less data, resulting in the $90\%$-quantile output only overshooting $67\%$ of the labels. Table~\ref{table:uncertainty_results} shows MPIW and PICP across different uncertainty levels $\alpha$. 
Our model consistently achieves better PICP values with prediction intervals roughly half as wide as those of \citeauthor{Lang2023}, whose prediction intervals moreover cover noticeably fewer labels than expected.

\subsubsection{Empirical Coverage across Target Bins}
\label{sec:res_target_bins}

\begin{figure}[t]
\centering
\includegraphics[width=\columnwidth]{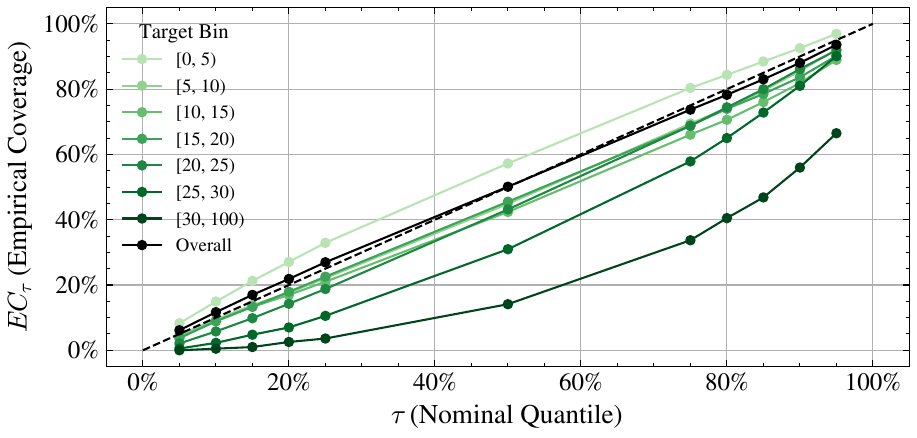}
\caption{Empirical Coverage vs Nominal Quantile for multiple target bins for our model.}
\label{fig:empirical_coverage_ours}
\end{figure}

\begin{figure*}[t]
    \centering
    \begin{subfigure}[t]{0.45\textwidth}
    \centering      \includegraphics[height=8cm]{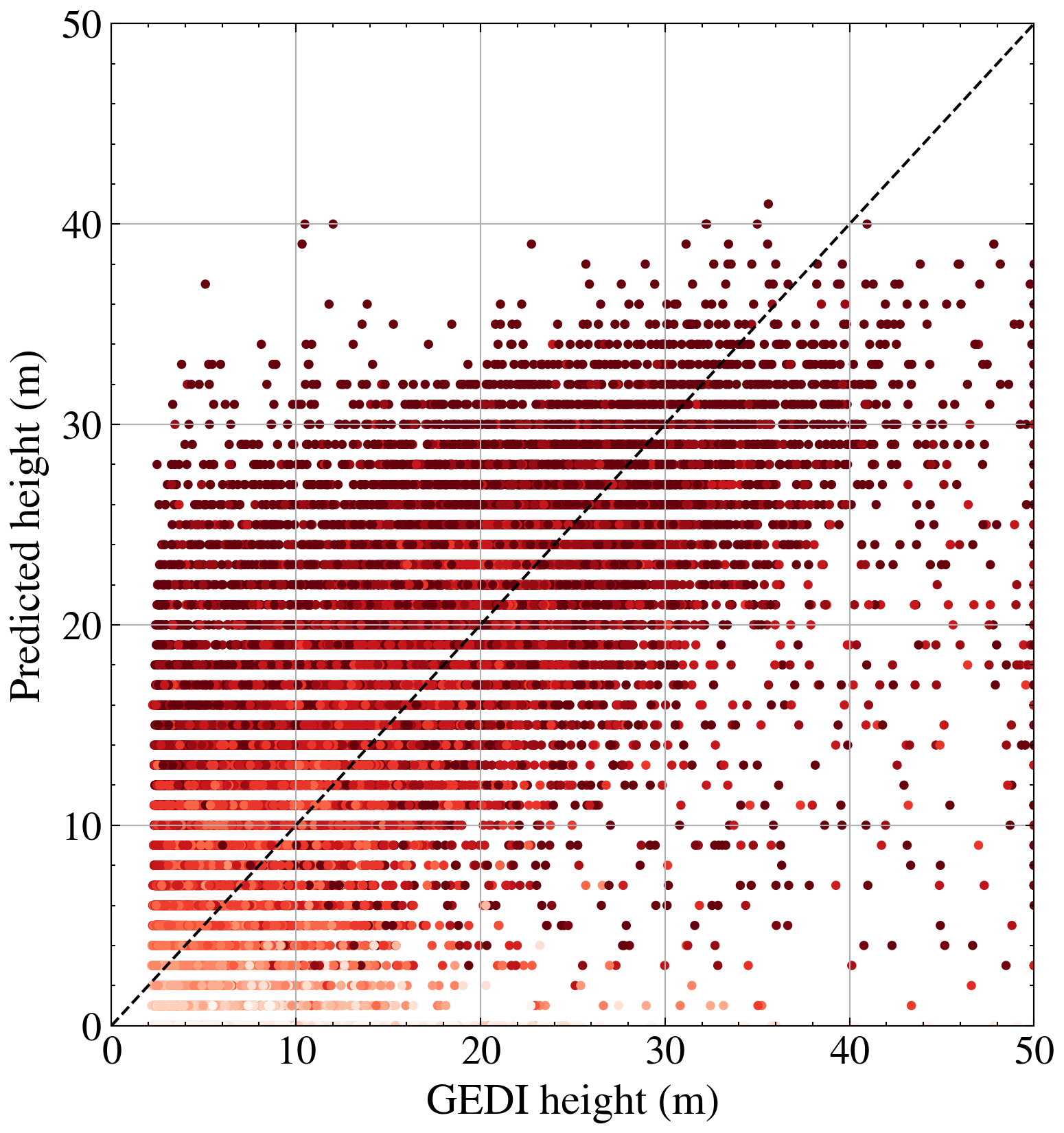}
    \end{subfigure}
    \hfill
    \begin{subfigure}[t]{0.49\textwidth}
    \centering  \includegraphics[height=8cm]{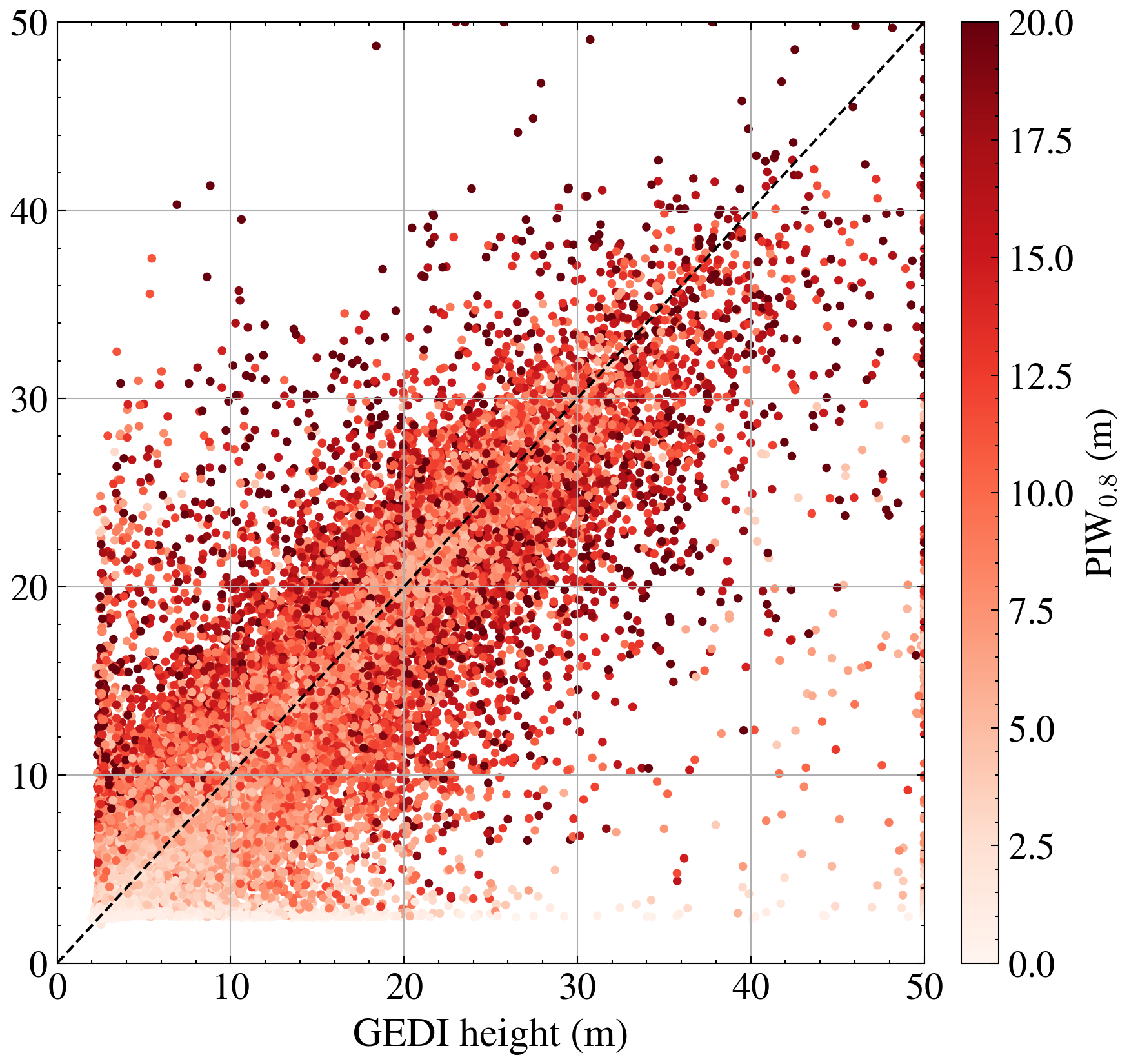}
    \end{subfigure}
    \hfill
    \vskip-0.2cm
    \caption{Scatter plot of GEDI height against predicted height with color dependent on the prediction interval width at level $\alpha = 0.8$. All heights are clipped into $[\SI{0}{\meter},\SI{50}{\meter}]$.
Left: \citeauthor{Lang2023}'s model. Right: Our model.}
    \label{fig:scatter_plot}
    \vspace{-0.2cm}
\end{figure*}

We evaluate the empirical coverage of predicted quantiles from our models across different target bins, results are depicted in Figure \ref{fig:empirical_coverage_ours}. For the $50\%$-quantile prediction, i.e. our point-estimator, we see the known pattern that tall trees are underpredicted, see also the scatter plot in Figure~\ref{fig:scatter_plot}. Additionally, here we see that all quantiles tend to overestimate small vegetation below \SI{5}{\meter} and underestimate taller trees. Note that for trees below \SI{25}{\meter}, the empirical coverage lies within a reasonable range of less than $5\%$ away from the expected value. For trees taller than $\SI{30}{\meter}$, even the $95\%$-quantile prediction covers less than $70\%$ of the trees. Note that correcting for these miscalibrated empirical coverages is non-trivial, as the miscalibration only manifests when predictions are grouped by target bins. When grouped by prediction bins instead, all groups are well-calibrated, compare Appendix~\ref{app:empirical_coverage_pred_bin}, Figure~\ref{fig:empirical_coverage_pred_bin}. 

\subsubsection{Prediction Interval Asymmetry}

\begin{figure}[h!]
\centering
\includegraphics[width=\columnwidth]{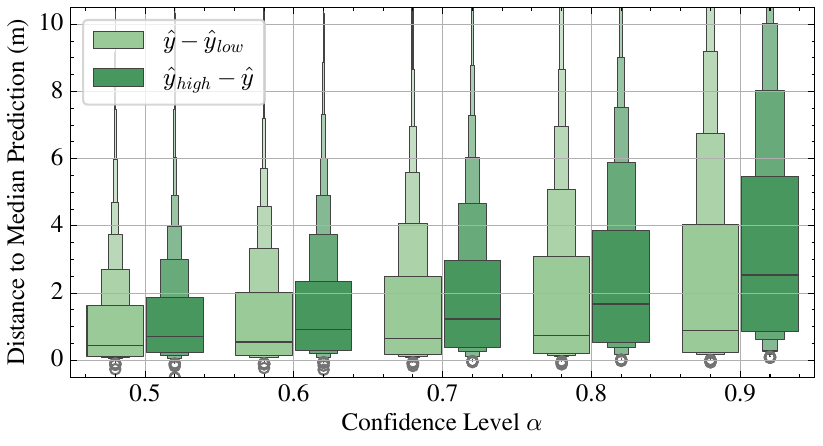}
\caption{Boxplot of distance of quantile predictions to point-estimate, i.e. median prediction. Per confidence level $\alpha \in [0.5,0.6,0.7,0.8,0.9]$, the prediction interval at level $\alpha$ is written as $PIW_\alpha = [\hat{y}_{low}, \hat{y}_{high}] \subset \mathbb{R}$, while the median prediction is denoted $\hat{y} \in \mathbb{R}$.}
\label{fig:quantile_dist_to_median}
\end{figure}

To test for asymmetry of the prediction intervals around the median prediction, Figure \ref{fig:quantile_dist_to_median} shows boxplots of the distances from the lower and upper edges of the prediction interval to the median prediction.

First, the distances increase with the confidence level $\alpha$, as expected, since higher-coverage prediction intervals are wider. Second, across all confidence levels, the upper edge is consistently further from the median than the lower edge, indicating a right-skewed predictive distribution. This asymmetry is particularly pronounced for the $90\%$ prediction interval, where the median distance of the upper edge (${\sim}\SI{2.54}{m}$) is roughly three times that of the lower edge (${\sim}\SI{0.8}{m}$). Notably, models that assume a symmetric distribution, such as the Gaussian regression approach of \citet{Lang2023}, are inherently unable to capture this behavior.

\subsubsection{Correlation of Prediction and Uncertainty}
\label{sec:res_high_pred_high_uncertainty}

\begin{figure}[t]
    \centering
    \includegraphics[width=0.99\columnwidth]{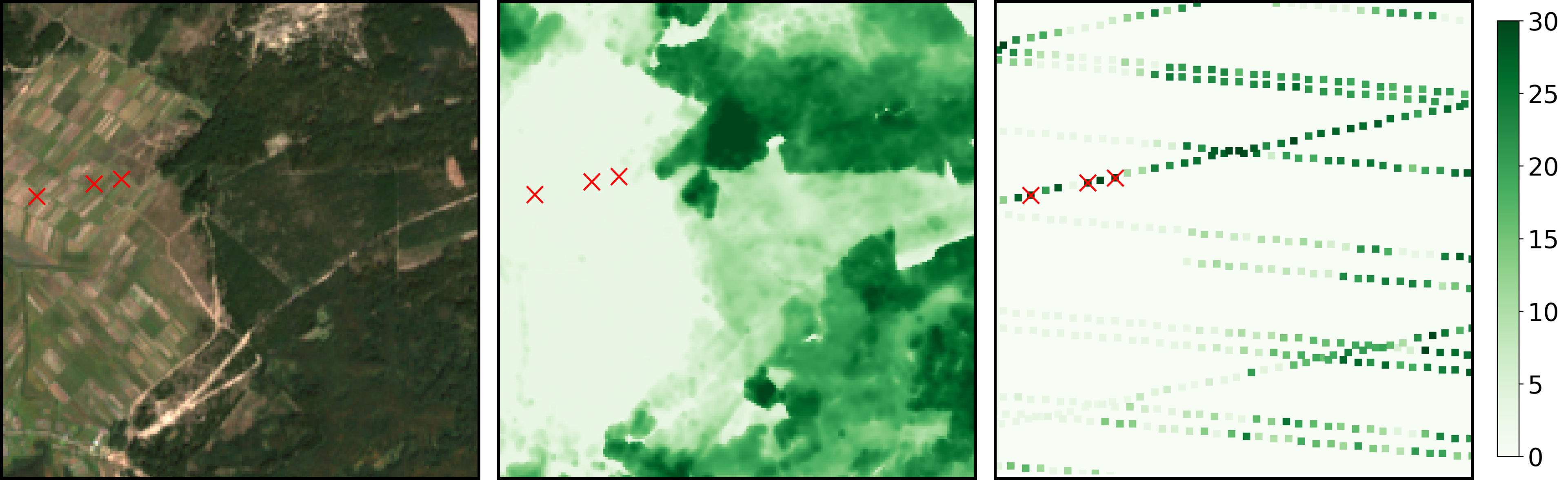}
    \caption{Noisy GEDI labels marked in red: $90\%$-quantile prediction below $\SI{10}{\meter}$ and GEDI labels above $\SI{30}{\meter}$. Left: Satellite image, Middle: $90\%$-quantile canopy height prediction, Right: GEDI labels}
    \label{fig:noisy_labels}
\end{figure}

Figure~\ref{fig:scatter_plot} compares GEDI reference height against predicted height with color-marked uncertainty, where a darker red indicates lower confidence. Note that \citeauthor{Lang2023}'s prediction are available in whole meters, leading to the horizontal stripes in the Figure.
Besides the observation that predictions of our model are more accurate overall, it is noticeable that for \citeauthor{Lang2023}'s model a small predicted uncertainty occurs only when the prediction is also low. In particular, the correlation between height prediction and uncertainty estimation is higher for \citeauthor{Lang2023} (Pearson correlation $0.85$ vs $0.72$ for our model), showing that uncertainty tends to rise with higher predictions.
For our model on contrast, there is a clear trend 
visible that points further away from the diagonal, i.e. points with higher errors, also have higher uncertainty, which is in line with what we expect from a good uncertainty model. 
However, there are also counterexamples. Note that there are some points, where the model is highly confident about the prediction of less than $\SI{10}{\meter}$, but the corresponding GEDI label is above $\SI{30}{\meter}$. An investigation of such situations indicate that these points might suffer from label noise, see Figure~\ref{fig:noisy_labels} for an example. The marked points lie on cropland, with a low prediction below $\SI{10}{\meter}$ of the $90\%$-conditional quantile, but have GEDI labels higher than $\SI{30}{\meter}$.  
Therefore, wrong predictions with high confidence might be an indicator of false labels. An approach to use this to detect and remove false labels from the dataset during training remains a subject of future work. Interestingly, there are much fewer points with a small GEDI label and a high prediction that the model is confident about.  

\subsubsection{Higher Uncertainty at Forest Borders}
\label{sec:res_forest_borders}

\begin{figure}[t]
\centering
\includegraphics[width=\columnwidth]{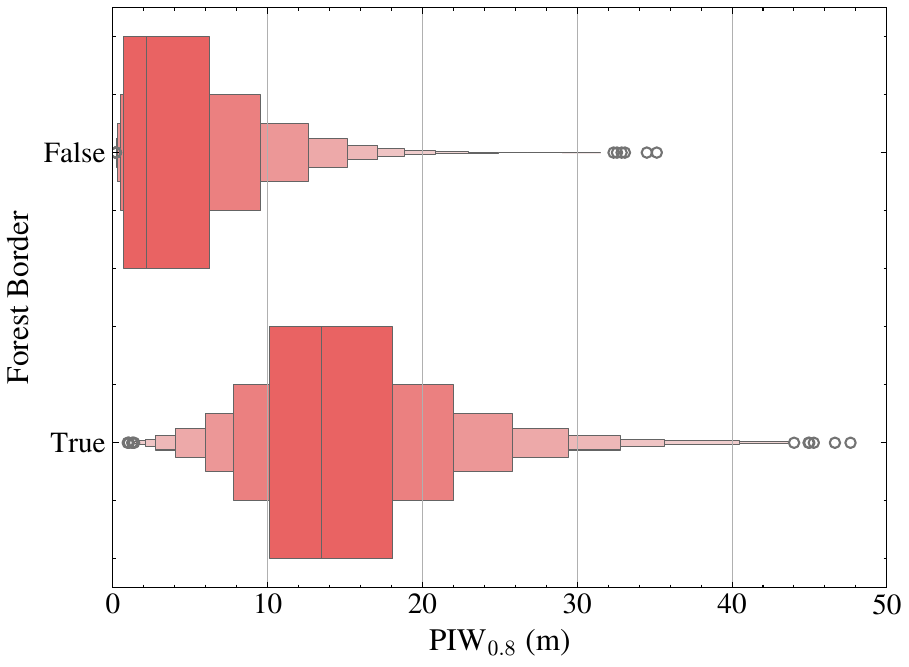}
\caption{Boxplot of Prediction Interval Width at level $\alpha = 0.8$ in dependence of being at a forest border. Note that less than $5\%$ of the pixels are located at forest borders.}
\label{fig:forest_border_boxplot}
\end{figure}

In this section, we investigate the predicted uncertainty at forest borders.
We define a pixel to lie at the border of a forest if the minimal and maximal predictions in the $3 \times 3$ surrounding of the pixel differ by more than $\SI{10}{m}$. 
Figure~\ref{fig:forest_border_boxplot} shows a boxplot of PIW between forest border pixels and non-forest border pixels. It is clearly visible that there is a huge difference in the PIW between pixels at the forest border and non-forest border pixels, meaning that, overall, our model is much less confident around forest borders. Additionally, the PICP for uncertainty level $\alpha = 0.8$ is about $78\%$ for non-forest border pixels and only $70\%$ for forest border pixels, indicating that uncertainty estimation is more complicated at forest borders. We hypothesize that this is at least partly a consequence of the geolocation error of GEDI measurements. Note that a geolocation offset at a forest border has a higher impact on the model than within a forest, where the spatial autocorrelation of tree heights is comparably high. Additionally, we observe an even higher uncertainty at forest borders, when the model is fine-tuned without the shift-resilient variant of the loss, see Appendix \ref{app:without_shift_loss} for details. 

\subsubsection{Mountaineous Areas}
\label{sec:res_mountains}

\begin{figure}[t]
\centering
\includegraphics[width=\columnwidth]{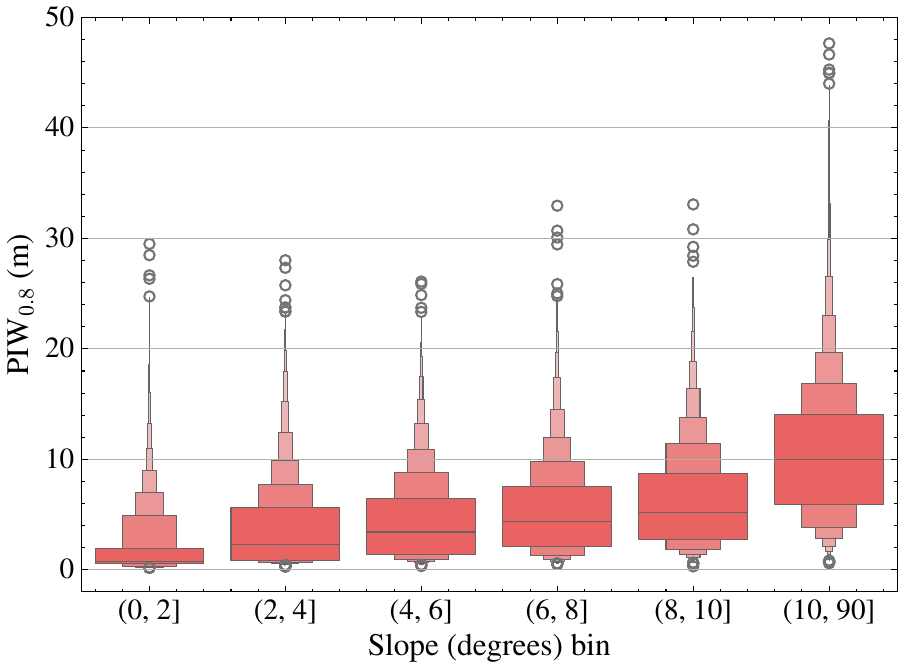}
\caption{Boxplot of Prediction Interval Width at level $\alpha = 0.8$ in dependence of average slope within a $\SI{140}{\meter}$ surrounding.}
\label{fig:slope_boxenplot}
\end{figure}

As GEDI's footprint size is approximately $25$m, measurements on steep slopes are overestimating true height, potentially having an estimated tree height for bare rocks. We therefore analyze the impact of slope on uncertainty by using the TandemX Edited Digital Elevation Model (EDEM) \citep{rs12233961} to calculate the average slope in degrees in a $\SI{140}{\meter}$ surrounding, roughly corresponding to a $3\times3$ kernel for the EDEM. Figure~\ref{fig:slope_boxenplot} shows a boxplot of PIW in dependence on 6 different slope bins. There is a trend that with increasing slope, our model predicts an increase in  PIW, as expected from higher label noise for steeper slopes for GEDI measurements \citep{fu2025accuracy}. Note that all points with a PIW above $\SI{35}{\meter}$ are in areas with a slope of at least $\SI{10}{\degree}$. These findings indicate the need of effectively dealing with GEDI measurements in steep areas to reduce noise in the labels. 

\begin{figure*}[t]
\centering
\includegraphics[width=1\textwidth]{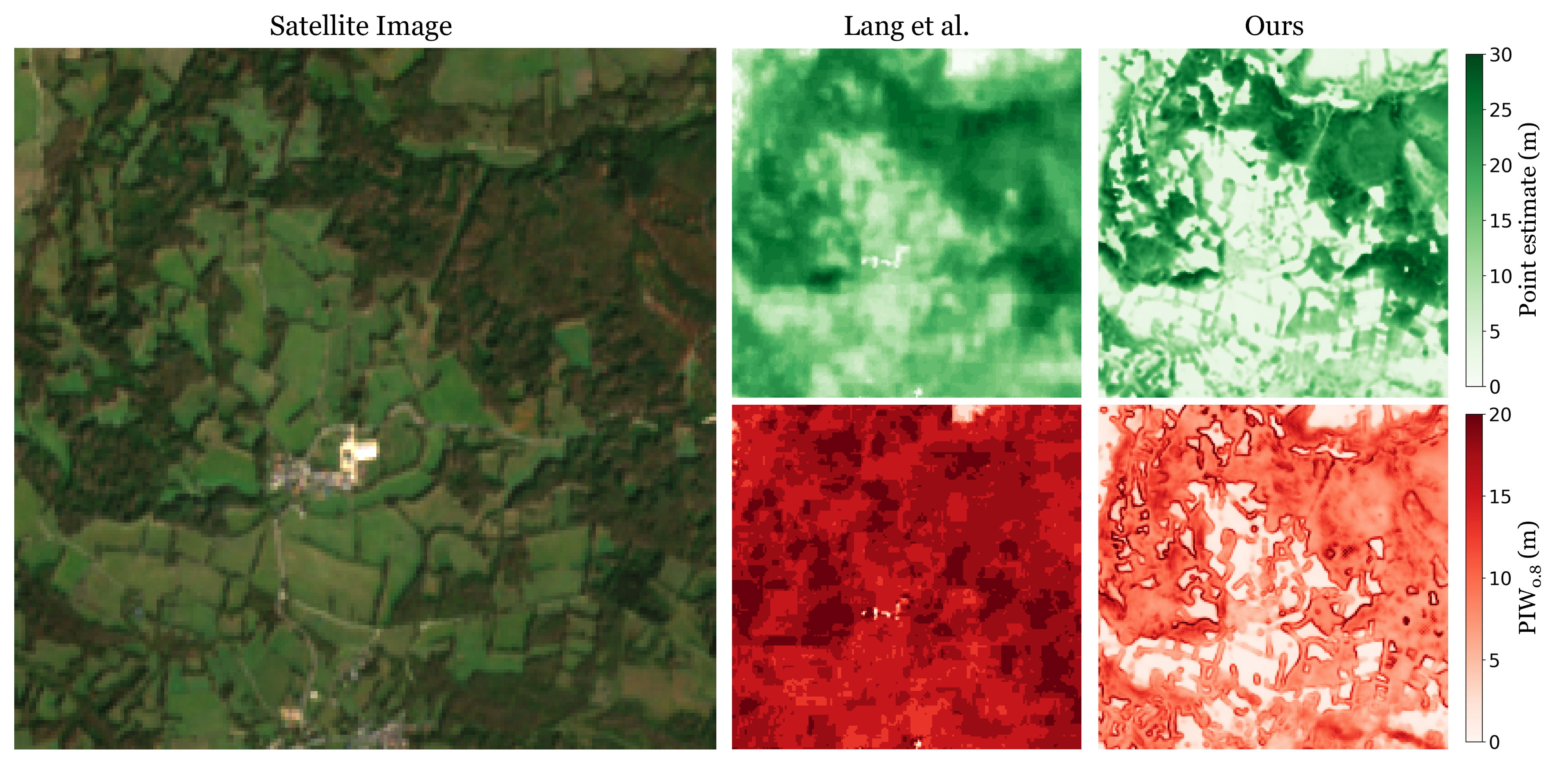}
\caption{Sample satellite image with point estimates and $80\%$ Prediction Interval Width of \citeauthor{Lang2023}'s model and our model. Left: Satellite image (yearly median composite). Top Right: Point-estimates of canopy height. Bottom Right: $80\%$ Prediction Inverval Width.}
\label{fig:qualitative_comparison}
\end{figure*}

\subsubsection{Qualitative Analysis} 
\label{sec:res_qualitative}

We qualitatively investigate uncertainty quantification based on one exemplary location, see Figure~\ref{fig:qualitative_comparison}. An immediate observation is that the predictions of \citeauthor{Lang2023}'s model are smoother than the predictions of our model. Except for this, the presented PIWs differ a lot. Overall, our model is more confident in its predictions compared to \citeauthor{Lang2023}'s model.
While the latter is less confident in areas with higher predictions, our model also shows confidence in some forest areas, see also Figure~\ref{fig:motivation} again. The highest uncertainty for our model arises in the highest areas of the forest, and at forest borders. Therefore, the shape of the forest can also be read off by the uncertainty map of our model. 
An increase in uncertainty at forest borders cannot be observed in the predictions from \citeauthor{Lang2023}'s model. Furthermore, note that our model is quite confident on grassland, where \citeauthor{Lang2023}'s model returns higher uncertainty. Note that the model from \citeauthor{Lang2023} is highly certain around the urban areas in the middle and the bottom of the image.  

\section{Conclusion}

In this study, we propose a lightweight modification on pretrained models to incorporate uncertainty estimation, with only fine-tuning the added uncertainty head of the model. For this, we suggest using quantile regression, as it is known to perform well on image-to-image tasks in biology, see \cite{Angelopoulos22}. 

This approach is experimentally tested in the application of tree height estimation, where a pretrained model from \cite{pauls2025} is used to add and fine-tune the uncertainty head. Uncertainty estimates are compared to an existing approach from \cite{Lang2023}. The results show that the uncertainty head can yield well-calibrated uncertainty quantification estimates after fine-tuning. 
The predicted quantiles are overall well-calibrated, although there are differences across target bins, as quantiles tend to underestimate tall trees. A general observation is that the model is less confident at forest borders and in steep terrain, hypothetically due to geolocation uncertainty in GEDI labels and label noise in mountainous areas. 
Another hypothesis is that wrong predictions with high confidence might be good indicators to detect faulty labels. A confirmation of this hypothesis and an eventual usage of detection and removal of faulty labels in training pipelines is a subject of future work. 

All in all, the results show that well-calibrated uncertainty quantification is possible in the application area of tree height prediction based on satellite images. This is indispensable for non-linear downstream applications such as biomass and carbon storage estimation, as policy-making requires knowledge of worst-case scenarios and risk assessment. 

\section*{Acknowledgements}

This work was supported via the AI4Forest project, which is funded by the German Federal Ministry of Education and Research (BMBF; grant number 01IS23025A) and the French National Research Agency (ANR). We also acknowledge the computational resources provided by the PALMA II cluster at the University of Münster (subsidized by the DFG; INST 211/667-1). 

\newpage
\bibliography{bibliography.bib}

\clearpage
\appendix
\thispagestyle{empty}

% Supplementary material: To improve readability, you must use a single-column format for the supplementary material.
\onecolumn
\aistatstitle{Canopy Tree Height Estimation Using Quantile Regression: Modeling and Evaluating Uncertainty in Remote Sensing \\
Supplementary Materials}

\section{Model Architecture}
\label{app:architecture}

The used architecture is depicted in Figure \ref{fig:unet3d_architecture}. Please note that the encoder and decoder originates in the 3d-Unet from \cite{pauls2025}, while one convolutional head is added to predict 10 quantiles. In particular, $C = 16$, $H = W = 256$, $C' = 64$ and $\mbox{out} = 11$ in our model. 

% ---- Define Weaker Colors ----
\definecolor{encoder_color}{HTML}{A5D6A7}
\definecolor{encoder_color_strong}{HTML}{81C784}
\definecolor{decoder_color}{HTML}{EF9A9A}
\definecolor{decoder_color_strong}{HTML}{E57373}
\definecolor{mha_color}{HTML}{90CAF9}
\definecolor{mha_color_strong}{HTML}{64B5F6}
\definecolor{ln_color}{HTML}{FFCC80}
\definecolor{ln_color_strong}{HTML}{FFB74D}
\definecolor{ffn_color}{HTML}{9FA8DA}
\definecolor{ffn_color_strong}{HTML}{7986CB}
\definecolor{other_color}{HTML}{E0E0E0}
\definecolor{other_color_strong}{HTML}{BDBDBD}
\definecolor{transformer_color}{HTML}{90A4AE}
\definecolor{transformer_color_strong}{HTML}{546E7A}
\definecolor{annotation_color}{HTML}{F06292}

% ---- Define Macros ----
\pgfmathsetmacro{\ssp}{0.2}
\pgfmathsetmacro{\msp}{0.4}
\pgfmathsetmacro{\bsp}{0.6}
\pgfmathsetmacro{\hsp}{1.1}

% ---- Define Styles ----
\tikzset{
    block/.style={
        draw, thick, rounded corners, minimum width=2cm, minimum height=0.6cm,
        draw=other_color_strong, fill=other_color
    },
    mha block/.style={block, draw=mha_color_strong, fill=mha_color},
    ffn block/.style={block, draw=ffn_color_strong, fill=ffn_color},
    ln block/.style={block, draw=ln_color_strong, fill=ln_color},
    main/.style={thick},
    main arrow/.style={thick, -{Latex}},
    bgbox/.style={draw, thick, inner sep=\ssp cm},
    annotation/.style={
        draw=annotation_color, dashed, fill=white,
        font=\fontsize{8}{8}\selectfont, text=annotation_color
    },
}

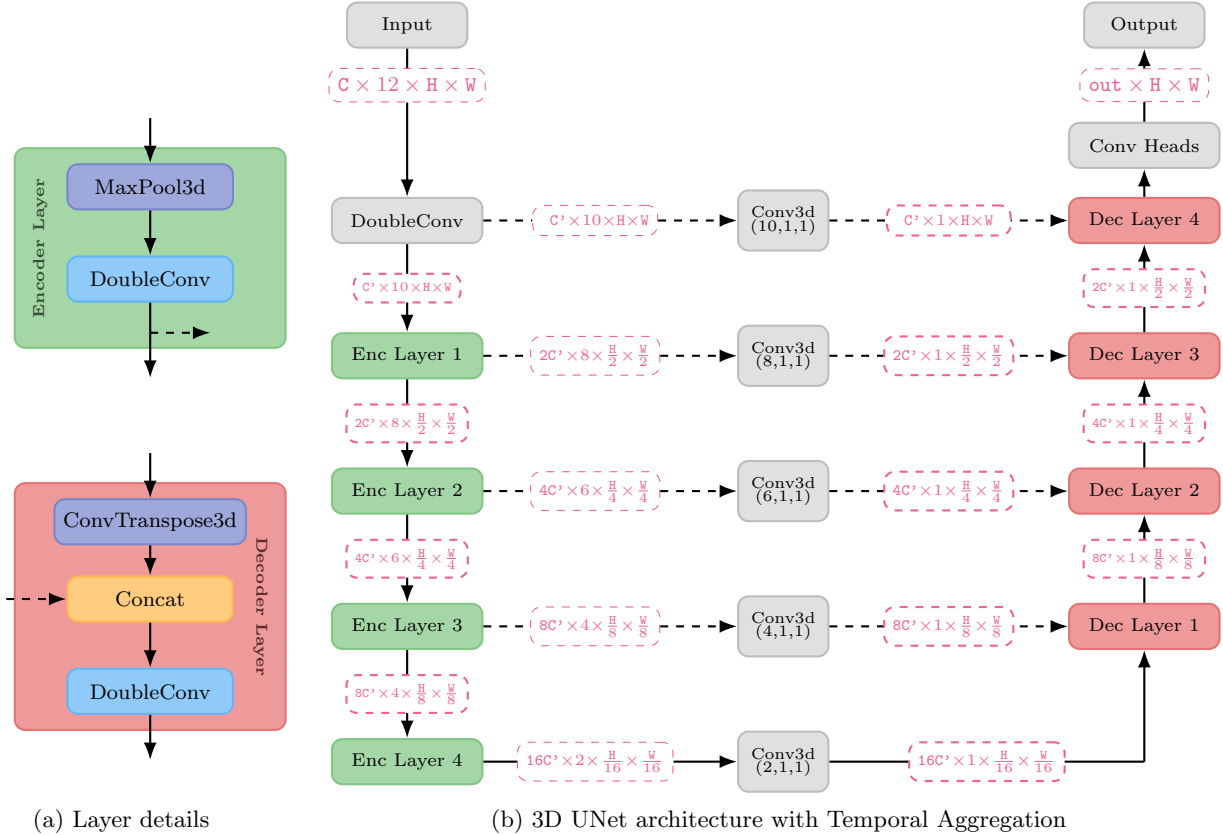
\begin{figure*}[h!]
    \centering
    %
    % ===== (a) Encoder & Decoder Layer Details =====
    %
    \begin{subfigure}[b]{0.18\textwidth}
        \centering
        {\begingroup\catcode`\^^M=5
            \begin{tikzpicture}[rounded corners=4pt, font=\fontsize{8}{8}\selectfont]
                % ---- Encoder Layer ----
                \coordinate (ei) at (0,0);
                \node[ffn block, minimum width=2.2cm, below=\bsp cm of ei] (mp) {MaxPool3d};
                \node[mha block, minimum width=2.2cm, below=\bsp cm of mp] (edc) {DoubleConv};
                \coordinate[below=\msp cm of edc] (ebranch);
                \coordinate[below=\bsp cm of ebranch] (eo);
                \coordinate[right=0.8cm of ebranch] (eskip);
                \draw[main arrow, dashed] (ebranch) -- (eskip);
                \draw[main arrow] (ei) -- (mp);
                \draw[main arrow] (mp) -- (edc);
                \draw[main arrow] (edc) -- (eo);
                \begin{scope}[on background layer]
                    \node[bgbox, draw=encoder_color_strong, fill=encoder_color,
                          minimum width=3.6cm,
                          fit={(mp)(edc)(eskip)}] (enc_bg) {};
                \end{scope}
                \node[rotate=90,
                      font=\fontsize{6}{6}\selectfont\bfseries,
                      text=encoder_color_strong!40!black,
                      anchor=center]
                      at ([xshift=3.5mm]enc_bg.west) {Encoder Layer};
                %
                % ---- Decoder Layer ----
                \coordinate[below=1.0cm of eo] (di);
                \node[ffn block, minimum width=2.2cm, below=\bsp cm of di] (ct) {ConvTranspose3d};
                \node[ln block, minimum width=2.2cm, below=\msp cm of ct] (cat) {Concat};
                \coordinate[left=0.8cm of cat] (dskip);
                \draw[main arrow, dashed] (dskip) -- (cat);
                \node[mha block, minimum width=2.2cm, below=\bsp cm of cat] (ddc) {DoubleConv};
                \coordinate[below=\bsp cm of ddc] (do);
                \draw[main arrow] (di) -- (ct);
                \draw[main arrow] (ct) -- (cat);
                \draw[main arrow] (cat) -- (ddc);
                \draw[main arrow] (ddc) -- (do);
                \begin{scope}[on background layer]
                    \node[bgbox, draw=decoder_color_strong, fill=decoder_color,
                          minimum width=3.6cm,
                          fit={(ct)(cat)(ddc)}] (dec_bg) {};
                \end{scope}
                \node[rotate=-90,
                      font=\fontsize{6}{6}\selectfont\bfseries,
                      text=decoder_color_strong!40!black,
                      anchor=center]
                      at ([xshift=-3.5mm]dec_bg.east) {Decoder Layer};
            \end{tikzpicture}
        \endgroup}
        \caption{Layer details}
        \label{fig:unet3d_a}
    \end{subfigure}
    \hfill
    %
    % ===== (b) Full Architecture with inline TA =====
    %
    \begin{subfigure}[b]{0.80\textwidth}
        \centering
        {\begingroup\catcode`\^^M=5
            \begin{tikzpicture}[rounded corners=4pt, font=\fontsize{7}{7}\selectfont]
                % ---- Column x-positions ----
                \pgfmathsetmacro{\colEnc}{0}
                \pgfmathsetmacro{\colPre}{2.5}
                \pgfmathsetmacro{\colTA}{5.0}
                \pgfmathsetmacro{\colDec}{9.8}
                % ---- Row y-positions (going down) ----
                \pgfmathsetmacro{\rA}{-1.8}
                \pgfmathsetmacro{\rB}{-3.6}
                \pgfmathsetmacro{\rC}{-5.4}
                \pgfmathsetmacro{\rD}{-7.2}
                \pgfmathsetmacro{\rE}{-9.0}

                % ======== INPUT BOX ========
                \node[block, minimum width=1.6cm] (input_box) at (\colEnc, 0.8) {Input};
                \node[annotation] (in_ann) at (\colEnc, 0)
                    {$\texttt{C} \times 12 \times \texttt{H} \times \texttt{W}$};

                \draw[main] (input_box) -- (in_ann);

                % ======== ENCODER (left column, downward) ========
                \node[block] (dc) at (\colEnc, \rA) {DoubleConv};
                \node[block, draw=encoder_color_strong, fill=encoder_color]
                    (e1) at (\colEnc, \rB) {Enc Layer 1};
                \node[block, draw=encoder_color_strong, fill=encoder_color]
                    (e2) at (\colEnc, \rC) {Enc Layer 2};
                \node[block, draw=encoder_color_strong, fill=encoder_color]
                    (e3) at (\colEnc, \rD) {Enc Layer 3};
                \node[block, draw=encoder_color_strong, fill=encoder_color]
                    (e4) at (\colEnc, \rE) {Enc Layer 4};

                % Encoder vertical flow with inline shape annotations
                \draw[main arrow] (in_ann) -- (dc);
                \draw[main arrow] (dc) --
                    node[midway, annotation, font=\fontsize{5}{5}\selectfont]
                    {$\texttt{C'} {\times} 10 {\times} \texttt{H} {\times} \texttt{W}$} (e1);
                \draw[main arrow] (e1) --
                    node[midway, annotation, font=\fontsize{5}{5}\selectfont]
                    {$\texttt{2C'} {\times} 8 {\times} \frac{\texttt{H}}{2} {\times} \frac{\texttt{W}}{2}$} (e2);
                \draw[main arrow] (e2) --
                    node[midway, annotation, font=\fontsize{5}{5}\selectfont]
                    {$\texttt{4C'} {\times} 6 {\times} \frac{\texttt{H}}{4} {\times} \frac{\texttt{W}}{4}$} (e3);
                \draw[main arrow] (e3) --
                    node[midway, annotation, font=\fontsize{5}{5}\selectfont]
                    {$\texttt{8C'} {\times} 4 {\times} \frac{\texttt{H}}{8} {\times} \frac{\texttt{W}}{8}$} (e4);

                % ======== PRE-TA SHAPES ========
                \node[annotation, font=\fontsize{6}{6}\selectfont] (pre0) at (\colPre, \rA)
                    {$\texttt{ C'} {\times} 10 {\times} \texttt{H} {\times} \texttt{W}$};
                \node[annotation, font=\fontsize{6}{6}\selectfont] (pre1) at (\colPre, \rB)
                    {$\texttt{2C'} {\times} 8 {\times} \frac{\texttt{H}}{2} {\times} \frac{\texttt{W}}{2}$};
                \node[annotation, font=\fontsize{6}{6}\selectfont] (pre2) at (\colPre, \rC)
                    {$\texttt{4C'} {\times} 6 {\times} \frac{\texttt{H}}{4} {\times} \frac{\texttt{W}}{4}$};
                \node[annotation, font=\fontsize{6}{6}\selectfont] (pre3) at (\colPre, \rD)
                    {$\texttt{8C'} {\times} 4 {\times} \frac{\texttt{H}}{8} {\times} \frac{\texttt{W}}{8}$};
                \node[annotation, font=\fontsize{6}{6}\selectfont] (pre4) at (\colPre, \rE)
                    {$\texttt{16C'} {\times} 2 {\times} \frac{\texttt{H}}{16} {\times} \frac{\texttt{W}}{16}$};

                % ======== TA CONV3D BLOCKS ========
                \node[block, align=center, minimum width=1.2cm, minimum height=0.8cm,
                      font=\fontsize{6}{6}\selectfont]
                    (ta0) at (\colTA, \rA) {Conv3d\\[-1pt]$(10{,}1{,}1)$};
                \node[block, align=center, minimum width=1.2cm, minimum height=0.8cm,
                      font=\fontsize{6}{6}\selectfont]
                    (ta1) at (\colTA, \rB) {Conv3d\\[-1pt]$(8{,}1{,}1)$};
                \node[block, align=center, minimum width=1.2cm, minimum height=0.8cm,
                      font=\fontsize{6}{6}\selectfont]
                    (ta2) at (\colTA, \rC) {Conv3d\\[-1pt]$(6{,}1{,}1)$};
                \node[block, align=center, minimum width=1.2cm, minimum height=0.8cm,
                      font=\fontsize{6}{6}\selectfont]
                    (ta3) at (\colTA, \rD) {Conv3d\\[-1pt]$(4{,}1{,}1)$};
                \node[block, align=center, minimum width=1.2cm, minimum height=0.8cm,
                      font=\fontsize{6}{6}\selectfont]
                    (ta4) at (\colTA, \rE) {Conv3d\\[-1pt]$(2{,}1{,}1)$};

                % ======== DECODER (right column, upward) ========
                \node[block, draw=decoder_color_strong, fill=decoder_color]
                    (d1) at (\colDec, \rD) {Dec Layer 1};
                \node[block, draw=decoder_color_strong, fill=decoder_color]
                    (d2) at (\colDec, \rC) {Dec Layer 2};
                \node[block, draw=decoder_color_strong, fill=decoder_color]
                    (d3) at (\colDec, \rB) {Dec Layer 3};
                \node[block, draw=decoder_color_strong, fill=decoder_color]
                    (d4) at (\colDec, \rA) {Dec Layer 4};

                % Conv Heads and Output — aligned with Input at y=0.8
                \node[block] (conv) at (\colDec, -0.8) {Conv Heads};
                \node[annotation] (out_ann) at (\colDec, 0)
                    {$\texttt{out} \times \texttt{H} \times \texttt{W}$};
                \node[block, minimum width=1.6cm] (output_box) at (\colDec, 0.8) {Output};

                % Decoder vertical flow with inline shape annotations (upward)
                \draw[main arrow] (d1) --
                    node[midway, annotation, font=\fontsize{5}{5}\selectfont]
                    {$\texttt{8C'} {\times} 1 {\times} \frac{\texttt{H}}{8} {\times} \frac{\texttt{W}}{8}$} (d2);
                \draw[main arrow] (d2) --
                    node[midway, annotation, font=\fontsize{5}{5}\selectfont]
                    {$\texttt{4C'} {\times} 1 {\times} \frac{\texttt{H}}{4} {\times} \frac{\texttt{W}}{4}$} (d3);
                \draw[main arrow] (d3) --
                    node[midway, annotation, font=\fontsize{5}{5}\selectfont]
                    {$\texttt{2C'} {\times} 1 {\times} \frac{\texttt{H}}{2} {\times} \frac{\texttt{W}}{2}$} (d4);
                \draw[main arrow] (d4) -- (conv);
                \draw[main] (conv) -- (out_ann);
                \draw[main arrow] (out_ann) -- (output_box);

                % ======== SKIP CONNECTIONS (dashed): Encoder → Pre → TA → Decoder ========
                % Post-TA shapes are now centered (midway) on the TA → Decoder segments

                % Level 0: DC → Dec4
                \draw[main, dashed] (dc.east) -- (pre0.west);
                \draw[main arrow, dashed] (pre0.east) -- (ta0.west);
                \draw[main arrow, dashed] (ta0.east) --
                    node[midway, annotation, font=\fontsize{6}{6}\selectfont]
                    {$\texttt{  C'} {\times} 1 {\times} \texttt{H} {\times} \texttt{W }$} (d4.west);

                % Level 1: E1 → Dec3
                \draw[main, dashed] (e1.east) -- (pre1.west);
                \draw[main arrow, dashed] (pre1.east) -- (ta1.west);
                \draw[main arrow, dashed] (ta1.east) --
                    node[midway, annotation, font=\fontsize{6}{6}\selectfont]
                    {$\texttt{2C'} {\times} 1 {\times} \frac{\texttt{H}}{2} {\times} \frac{\texttt{W}}{2}$} (d3.west);

                % Level 2: E2 → Dec2
                \draw[main, dashed] (e2.east) -- (pre2.west);
                \draw[main arrow, dashed] (pre2.east) -- (ta2.west);
                \draw[main arrow, dashed] (ta2.east) --
                    node[midway, annotation, font=\fontsize{6}{6}\selectfont]
                    {$\texttt{4C'} {\times} 1 {\times} \frac{\texttt{H}}{4} {\times} \frac{\texttt{W}}{4}$} (d2.west);

                % Level 3: E3 → Dec1 (skip)
                \draw[main, dashed] (e3.east) -- (pre3.west);
                \draw[main arrow, dashed] (pre3.east) -- (ta3.west);
                \draw[main arrow, dashed] (ta3.east) --
                    node[midway, annotation, font=\fontsize{6}{6}\selectfont]
                    {$\texttt{8C'} {\times} 1 {\times} \frac{\texttt{H}}{8} {\times} \frac{\texttt{W}}{8}$} (d1.west);

                % ======== BOTTLENECK (solid): E4 → TA → Dec1 ========
                \draw[main] (e4.east) -- (pre4.west);
                \draw[main arrow] (pre4.east) -- (ta4.west);
                % Horizontal segment with centered annotation, then vertical to Dec1
                \coordinate (bot_corner) at (\colDec, \rE);
                \draw[main] (ta4.east) --
                    node[midway, annotation, font=\fontsize{6}{6}\selectfont]
                    {$\texttt{16C'} {\times} 1 {\times} \frac{\texttt{H}}{16} {\times} \frac{\texttt{W}}{16}$}
                    (bot_corner);
                \draw[main arrow] (bot_corner) -- (d1.south);

            \end{tikzpicture}
        \endgroup}
        \caption{3D UNet architecture with Temporal Aggregation}
        \label{fig:unet3d_arch}
    \end{subfigure}

    \caption{Architecture of the 3D UNet.
    (a)~Each Encoder Layer applies MaxPool3d for $2{\times}$ spatial downsampling followed by a DoubleConv block (two Conv3d\,+\,GroupNorm\,+\,ReLU); the initial DoubleConv omits the MaxPool.
    Each Decoder Layer applies ConvTranspose3d for $2{\times}$ spatial upsampling, concatenates skip features (dashed), then applies a DoubleConv block.
    (b)~Overall architecture. Encoder features pass through Temporal Aggregation Conv3d layers with temporal-only kernels $(T{,}1{,}1)$ that collapse the time dimension to~1 before entering the decoder as skip connections.
    The solid path at the bottom shows the bottleneck flow.
    Tensor shapes are shown in magenta as $C \times T \times H \times W$.}
    \label{fig:unet3d_architecture}
\end{figure*}

\section{Ablation study with Gaussian Regression and Log-Gaussian Regression}
\label{app:ablation_gaussian}

As an ablation study, we compare our quantile regression approach against two alternative uncertainty estimation methods: Gaussian Regression and Log-Gaussian Regression. In Gaussian Regression, the model predicts, for each pixel, a mean $\mu$ and a log-variance $\log \sigma^2$, and is trained by minimizing the Gaussian negative log-likelihood:
$$\mathcal{L}(\mu, \sigma^2, y) = \frac{1}{2} \log \sigma^2 + \frac{(y - \mu)^2}{2\sigma^2},$$
following the same formulation as \citet{Lang2023}. Compared to our architecture, this requires only one additional output channel (for $\log \sigma^2$) instead of ten (for the quantile predictions), resulting in comparable computational costs.
A limitation of Gaussian Regression is that the symmetric Gaussian distribution assigns non-zero probability to negative tree heights. To address this, we additionally consider Log-Gaussian Regression, which replaces the Gaussian likelihood with a log-normal likelihood. In this formulation, the model predicts the parameters of a Gaussian distribution in log-space, ensuring that the implied distribution over tree heights has support strictly on the positive real line. The network architecture remains unchanged from Gaussian Regression. All models are trained with and without the Shift-Resilient Loss (see Section \ref{sec:shift_resilient_loss}).

Table~\ref{tab:point_results_ablation} and Table \ref{tab:uncertainty_results_ablation} report the same point estimation and uncertainty metrics as in the main paper for all models tested.

\begin{table}[h]
    \centering
    \caption{Results on point-estimators, measured using Mean Squared Error (MSE, \si{\meter\squared}), Mean Absolute Error (MAE, \si{\meter}), $R^2$ and Empirical Coverage (EC).}
    \label{tab:point_results_ablation}
\begin{tabular}{l|cccc}
\toprule
Model & MSE $\downarrow$ & MAE $\downarrow$ & R2 $\uparrow$ & $EC_{0.5}$ \\
\midrule
Lang et. al. & 38.32 & 4.25 & 0.55 & 0.47 \\
Pauls et. al. & 20.64 & 1.90 & 0.76 & 0.58 \\
Gaussian w/o Shift & 22.11 & 2.17 & 0.74 & 0.58 \\
Gaussian w/ Shift & 22.54 & 2.17 & 0.74 & 0.58 \\
Log Gaussian w/o Shift & 21.86 & 2.05 & 0.74 & 0.54 \\
Log Gaussian w/ Shift & 21.86 & 2.04 & 0.74 & 0.54 \\
Ours wo/ Shift & 20.34 & 1.88 & 0.76 & 0.50 \\
Ours & 20.81 & 1.90 & 0.76 & 0.50 \\
\bottomrule
\end{tabular}
\end{table}

\begin{table}[h]
\centering
\caption{Comparison of Mean Prediction Interval Width (MPIW) and Prediction Interval Coverage Probability (PICP) for multiple uncertainty levels $\alpha$.}
    \label{tab:uncertainty_results_ablation}
\begin{tabular}{r|ccccc|ccccc}
\toprule
Metric & \multicolumn{5}{c|}{MPIW} & \multicolumn{5}{c}{PICP} \\
 $\alpha$ & $0.5$ & $0.6$ & $0.7$ & $0.8$ & $0.9$ & $0.5$ & $0.6$ & $0.7$ & $0.8$ & $0.9$ \\
\midrule
Lang et. al. & 5.54 & 6.87 & 8.39 & 10.22 & 12.76 & 0.37 & 0.45 & 0.52 & 0.58 & 0.64 \\
Gaussian w/o Shift & 3.93 & 4.91 & 6.04 & 7.47 & 9.59 & 0.70 & 0.78 & 0.85 & 0.90 & 0.94 \\
Gaussian w/ Shift & 3.79 & 4.72 & 5.82 & 7.19 & 9.23 & 0.66 & 0.75 & 0.82 & 0.88 & 0.93 \\
Log Gaussian w/o Shift & 3.67 & 4.62 & 5.76 & 7.24 & 9.61 & 0.63 & 0.74 & 0.82 & 0.89 & 0.94 \\
Log Gaussian w/ Shift & 3.35 & 4.22 & 5.25 & 6.59 & 8.73 & 0.58 & 0.69 & 0.79 & 0.86 & 0.92 \\
Ours w/o Shift & 2.69 & 3.34 & 4.15 & 5.25 & 7.10 & 0.49 & 0.59 & 0.69 & 0.79 & 0.90 \\
Ours & 2.43 & 3.01 & 3.72 & 4.68 & 6.34 & 0.47 & 0.56 & 0.66 & 0.76 & 0.87 \\
\bottomrule
\end{tabular}
\end{table}

\section{Computational Overhead}
\label{app:computation_overhead}
In this section we provide the results of a comparison between our proposed method and the one, which we build upon \citep{pauls2025}. It can be seen that the additional uncertainty head adds less than $10\%$ of extra time and less than $13 \%$ additional VRAM usage during inference.

\begin{table}[h]
\caption{VRAM usage and compute time during inference and training for a single input, FLOPs and number of parameters from our proposed method and \cite{pauls2025}, both with either frozen backbone and trainable head (Ours and \citeauthor{pauls2025}$_{\textit{head\_only}}$) or fully trainable (Ours$_{\textit{full}}$ and \citeauthor{pauls2025}$_{\textit{full}}$ ).}
\label{tab:resource_table}
\begin{center}
\begin{tabular}{l|cc|cc|cc}
\toprule
 & \multicolumn{2}{c|}{VRAM} 
 & \multicolumn{2}{c|}{Time (ms)} 
 & FLOPS & Params \\
Model 
& Inference & Training 
& Inference  & Training
& (G) & (M) \\
\midrule
Ours 
  & 1668MiB & 1672MiB 
  & 2.79 $\pm$ 0.72 & 3.82 $\pm$ 0.18 
  & 430.531 & 85.0659 \\
Ours$_{\textit{full}}$  
  & 1668MiB & 3810MiB 
  & 2.79 $\pm$ 0.72 & 7.54 $\pm$ 0.25 
  & 430.531 & 85.0659 \\
\citeauthor{pauls2025}$_{\textit{head\_only}}$ 
  & 1476MiB & 1478MiB 
  &2.55 $\pm$ 0.97 & 3.00 $\pm$ 0.17 
  & 425.623 & 85.0653 \\
\citeauthor{pauls2025}$_{\textit{full}}$ 
  & 1476MiB & 3900MiB 
  & 2.55 $\pm$ 0.97 & 6.77 $\pm$ 0.26 
  & 425.623 & 85.0653 \\
\bottomrule
\end{tabular}
\end{center}
\end{table}

\section{Empirical Coverage across Prediction Bins}
\label{app:empirical_coverage_pred_bin}
\begin{figure*}[h!]
\centering
\includegraphics[width=0.475\columnwidth]{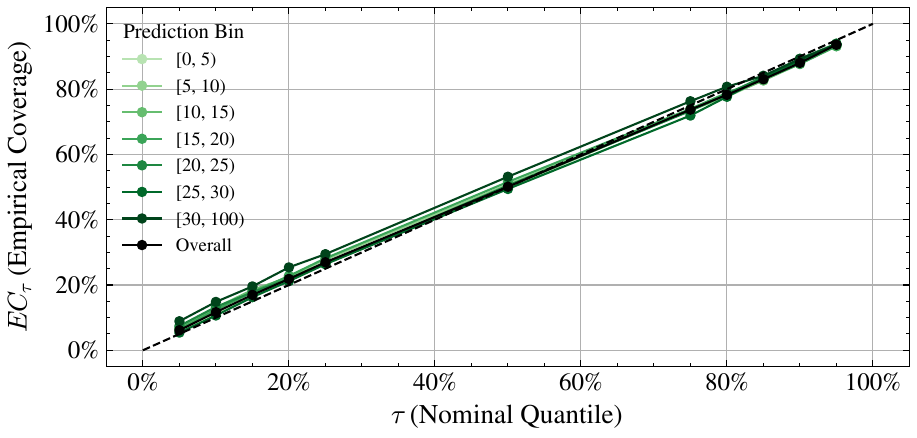}
\caption{Empirical Coverage vs Nominal Quantile for multiple prediction bins for our model. Across all prediction bins, the conditional quantiles are well-calibrated.}
\label{fig:empirical_coverage_pred_bin}
\end{figure*}

The empirical coverage of quantile predictions across predictions bins is depicted in Figure \ref{fig:empirical_coverage_pred_bin}. 

\section{Visualizations for our Model without Shifted Loss}
\label{app:without_shift_loss}

\begin{figure}[h!]
\centering
    \begin{subfigure}{0.49\textwidth}
        \centering      \includegraphics[width=0.95\linewidth]{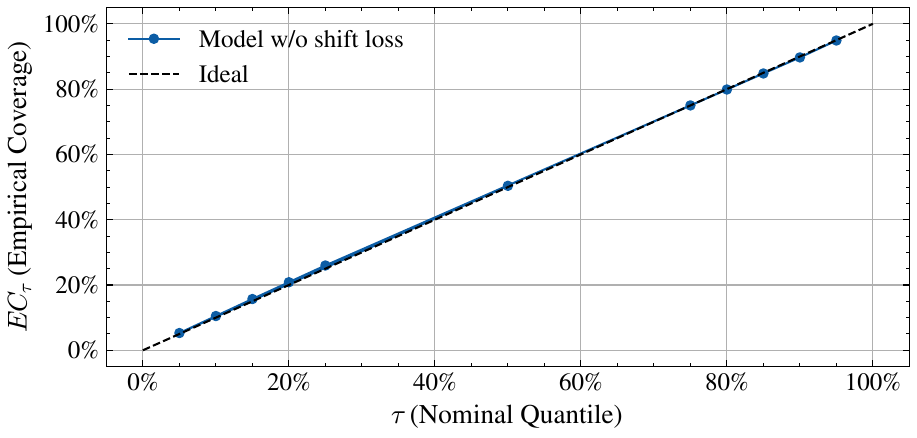}
    \end{subfigure}
    \hfill
    \begin{subfigure}{0.49\textwidth}
        \centering  \includegraphics[width=0.95\linewidth]{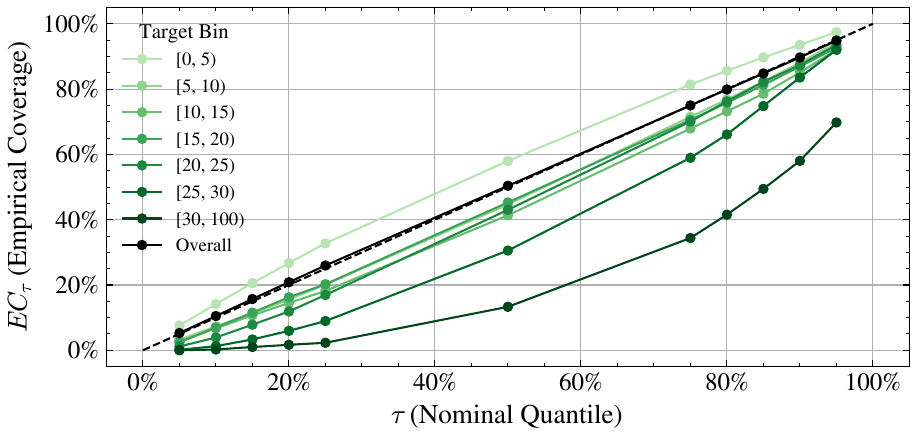}
    \end{subfigure}
\caption{Empirical Coverage vs Nominal Quantile for our model trained without shift loss.}
\label{fig:empirical_coverage_wo_shift}
\end{figure}

\begin{figure}[h!]
\centering
    \begin{subfigure}[t]{0.49\textwidth}
        \centering      \includegraphics[width=0.95\linewidth]{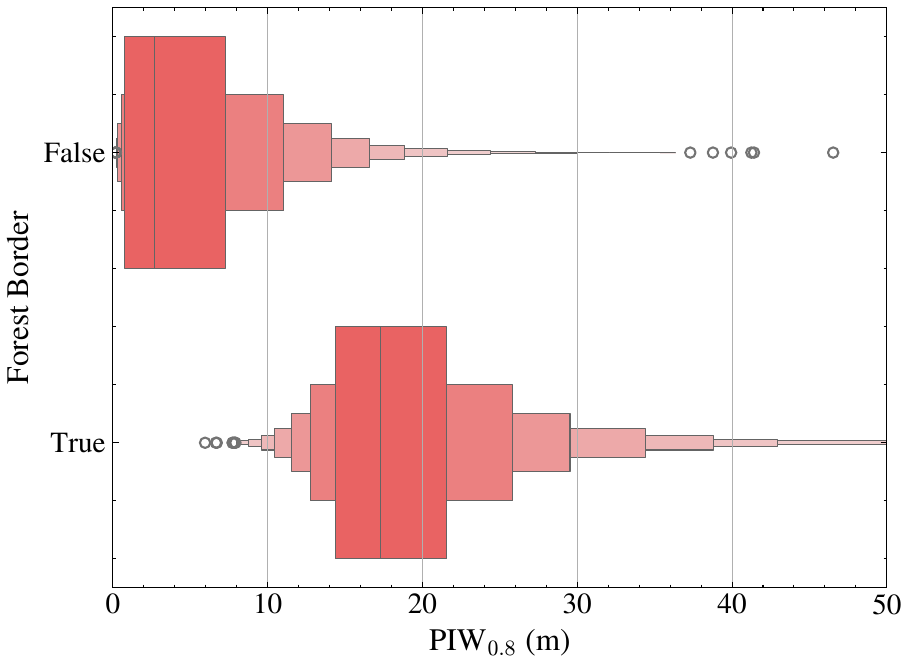}
    \end{subfigure}
    \hfill
    \begin{subfigure}[t]{0.49\textwidth}
        \centering  \includegraphics[width=0.95\linewidth]{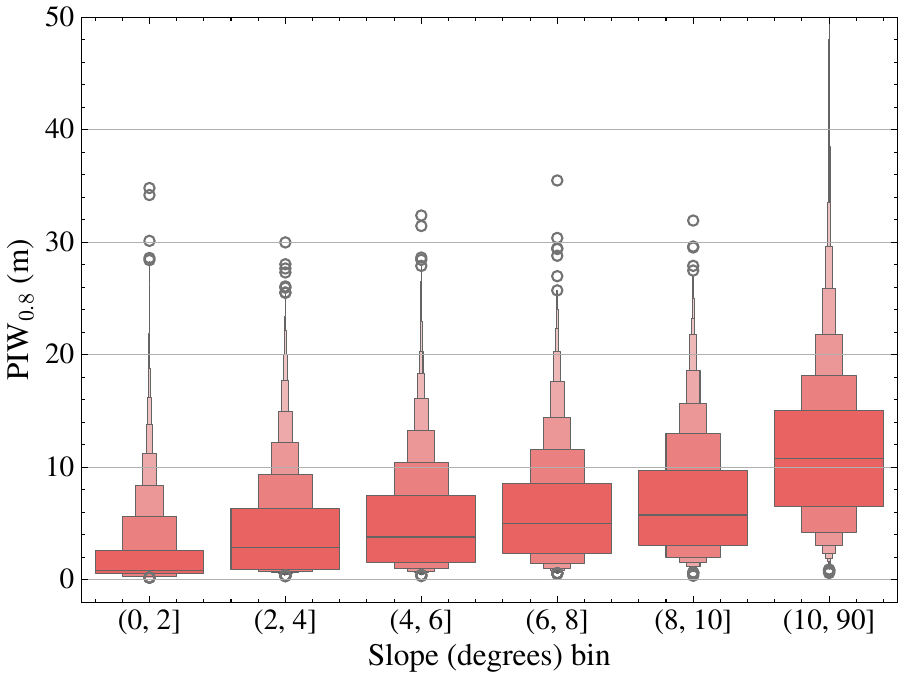}
    \end{subfigure}
\caption{Boxplot of Prediction Interval Width at level $\alpha = 0.8$ for model trained without shift loss in dependence of either Left: being at forest border or Right: average slope within a $\SI{140}{\meter}$ square. Note that the model without shift loss is more uncertain at forest borders compared to the model with shift loss (compare Figures \ref{fig:forest_border_boxplot} and \ref{fig:slope_boxenplot}).}
\label{fig:boxenplot_wo_shift}
\end{figure}

\begin{figure*}[h!]
\centering
\includegraphics[width=0.89\textwidth]{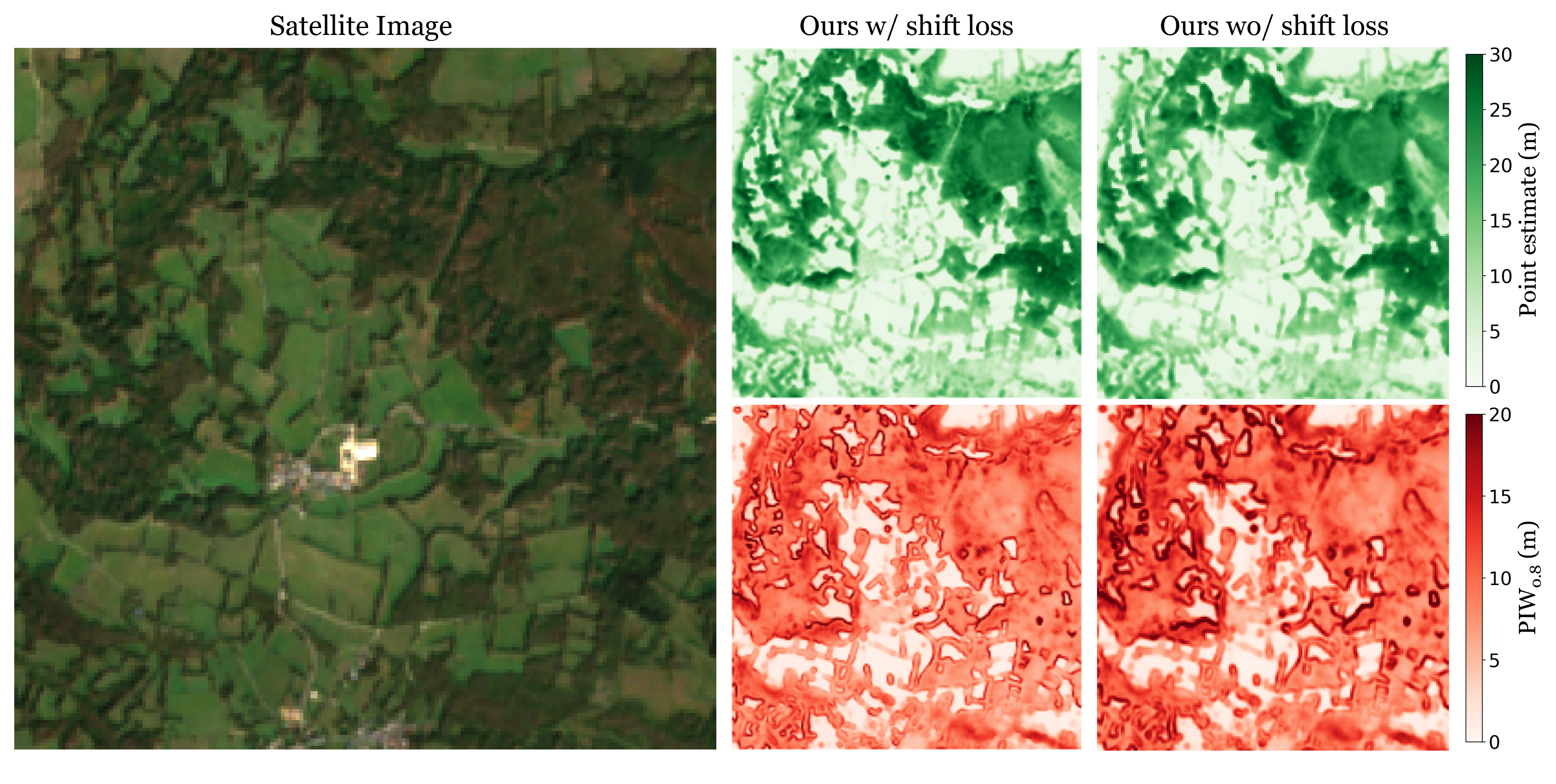}
\caption{Sample satellite image with point estimates and $80\%$ Prediction Interval Width of our model, trained with or without shift loss. Left: Satellite image (yearly median composite). Top Right: Point-estimates of canopy height. Bottom Right: $80\%$ Prediction Inverval Width. It is clearly visible that the shift loss reduces uncertainty at forest borders.}
\label{fig:qualitative_comparison_wo_shift}
\end{figure*}

Figure~\ref{fig:empirical_coverage_wo_shift} shows the EC for different quantiles on the left and split up by target bins on the right side, Figure~\ref{fig:boxenplot_wo_shift} shows the boxenplots for forest borders and different slope values and Figure~\ref{fig:qualitative_comparison_wo_shift} shows a visual comparison between training our model with and without shifted loss, showing that the model without shift loss is much less confident at forest borders. 

\end{document}